\documentclass[journal]{IEEEtran}
\usepackage{amsmath,amssymb}
\usepackage{graphicx}
\usepackage{booktabs}
\usepackage{multirow}
\usepackage{cite}
\usepackage{url}
\usepackage{xcolor}
\usepackage{float}

\begin{document}

\title{A Comprehensive Survey of Wireless Foundation Models for AI-Native 6G Networks}

\author{Naveed Khan,
Besan~Al~Sbeihi,
Maryam~Alshehhi,~\IEEEmembership{Student Member,~IEEE},
and
Nasir~Saeed,~\IEEEmembership{Senior Member,~IEEE}
\thanks{The authors are with the Department of Electrical and Communication Engineering, College of Engineering,
United Arab Emirates University (UAEU), Al Ain, UAE (e-mail: mr.nasir.saeed@ieee.org).}}


\maketitle


\begin{abstract}

Foundation models are emerging as a transformative paradigm for AI-native sixth-generation (6G) wireless networks by enabling scalable, transferable, and data-efficient intelligence across diverse communication tasks. Unlike conventional deep learning models that are trained for individual applications, wireless foundation models (WFMs) learn generalized representations from large-scale heterogeneous wireless data and can be efficiently adapted to communication, sensing, localization, and network optimization tasks with minimal task-specific supervision. Despite rapid progress, current research remains fragmented across architectures, training paradigms, and application domains, with no unified survey dedicated to the design, learning, and deployment of WFMs.
This survey presents a comprehensive and unified review of wireless foundation models. We first establish the fundamental concepts of WFMs and introduce a taxonomy that organizes the field according to model architectures, pre-training paradigms, and applications. We then review representative architectures, self-supervised pre-training strategies, parameter-efficient adaptation methods, datasets, benchmarks, and evaluation methodologies, highlighting their roles in enabling transferable wireless intelligence. Furthermore, we examine emerging applications spanning physical-layer signal processing, network intelligence, and cross-layer optimization, and discuss the key challenges of data availability, generalization, interpretability, efficient edge deployment, and standardization. Finally, we outline future research directions toward scalable, trustworthy, and general-purpose wireless intelligence for AI-native 6G networks. This survey provides a comprehensive reference for researchers and practitioners developing next-generation intelligent wireless systems.

\end{abstract}


\section{Introduction}
Wireless communication systems are evolving from model-driven signal processing toward data-driven intelligence to meet the stringent performance, scalability, and adaptability requirements of AI-native sixth-generation (6G) networks \cite{cui2025overview}. For decades, analytical channel models, estimation theory, optimization techniques, and signal processing algorithms have successfully supported wireless tasks such as channel estimation, signal detection, beamforming, resource allocation, and network management \cite{liu2024survey}. However, emerging technologies, including massive multiple-input multiple-output (MIMO), ultra-dense heterogeneous networks, integrated sensing and communication (ISAC), semantic communications, and AI-native 6G systems, have introduced levels of complexity that increasingly challenge conventional analytical models \cite{Dai2020DLSurvey,Yu2022RoleDL,Jiang2026ComprehensiveSurvey}.

Deep learning has significantly advanced wireless communications by learning complex signal representations directly from data, with early successes in physical-layer tasks such as channel estimation, signal detection, and beamforming \cite{Dai2020DLSurvey}. Its use has subsequently expanded to localization, modulation recognition, intelligent wireless receivers, and network-level resource optimization \cite{Yu2022RoleDL,Doha2025DLReceivers}. Nevertheless, most existing learning-based approaches remain task-specific, requiring separate models and training procedures for individual communication problems. Moreover, their dependence on large labeled datasets and environment-specific training can make adaptation to new deployment conditions costly \cite{Nguyen2021TransferLearning}. Models trained under particular channel, mobility, or hardware conditions may also generalize poorly when the underlying data distribution changes \cite{Akrout2023DomainGeneralization,Liu2021OOD}. As wireless systems evolve toward increasingly dynamic and heterogeneous environments, repeatedly collecting labeled data, retraining models, and maintaining independent learning pipelines for individual tasks becomes inefficient and difficult to scale.

These limitations have motivated a transition from task-specific learning toward foundation models. Rather than optimizing separate models for individual tasks, foundation models are pre-trained on large-scale and heterogeneous datasets, often using self-supervised or unsupervised objectives, to learn transferable representations that can be efficiently adapted to multiple downstream applications \cite{zhou2025comprehensive}. This paradigm has already reshaped natural language processing and computer vision through improved data efficiency, generalization, and knowledge transfer, and is now attracting growing interest in wireless communications \cite{abdelraouf2026paradigm}. 

Wireless systems are particularly well suited to this paradigm because they generate diverse and complementary data modalities. Channel-centric measurements, such as channel state information (CSI) and channel impulse responses (CIR), provide rich representations of propagation characteristics and spatial structure \cite{Buffelli2025FoundationCommunicationSystems}. At the signal level, in-phase and quadrature (IQ) samples, spectrograms, and other radio-frequency (RF) observations enable foundation models to learn reusable representations directly from raw or transformed wireless signals \cite{Alikhani2025LWM,Yang2025WirelessGPT}. More recent approaches further incorporate heterogeneous sensing and communication modalities to capture complementary information across wireless environments \cite{Aboulfotouh2025WavesFM}. The integration of channel, signal, sensing, and network-level observations therefore provides a natural basis for multimodal representation learning and the development of general-purpose wireless foundation models \cite{Zhang2026MultiModalSurvey}.

\begin{figure*}[t]
    \centering
    \includegraphics[width=0.95\textwidth]{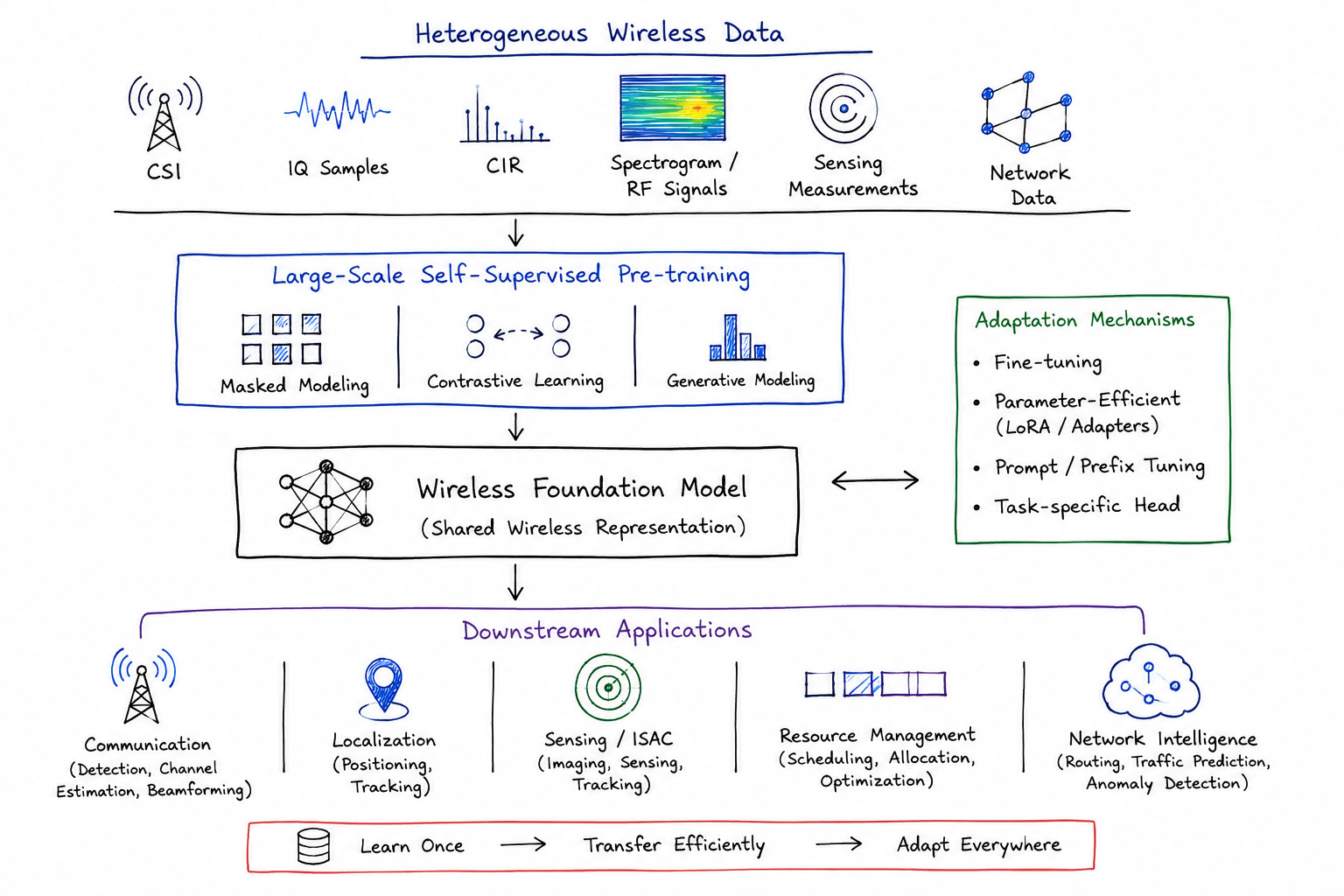}
   \caption{Conceptual workflow of a wireless foundation model. Large-scale self-supervised pre-training learns a shared wireless representation from heterogeneous wireless data, which is subsequently adapted to diverse downstream communication, sensing, localization, and network intelligence tasks through lightweight adaptation mechanisms.}
    \label{fig:wfm_working_mechanism}
\end{figure*}

Wireless foundation models (WFMs) have consequently emerged as a promising paradigm for enabling scalable and general-purpose wireless intelligence. Rather than developing independent models for communication, sensing, localization, and network optimization, a single pre-trained backbone can be efficiently adapted to multiple downstream tasks through fine-tuning, parameter-efficient adaptation, prompt tuning, or in-context learning \cite{khan2026model}. Recent models, including WirelessGPT, Large Wireless Model (LWM), WavesFM, and emerging multimodal WFMs, demonstrate that large-scale pre-training substantially improves knowledge transfer, parameter efficiency, and cross-domain generalization while reducing dependence on large labeled datasets \cite{wang2026hierarchical}. Figure.~\ref{fig:wfm_working_mechanism} summarizes the fundamental learning paradigm of WFMs. Unlike conventional deep learning, where separate neural networks are independently trained for individual communication tasks, WFMs decouple large-scale representation learning from task-specific optimization. During pre-training, heterogeneous wireless data including CSI, IQ samples, CIRs, RF signals, sensing measurements, and network observations are used to learn a shared representation that captures common propagation characteristics across diverse wireless environments. The resulting backbone is subsequently adapted to multiple downstream tasks through lightweight techniques such as fine-tuning, parameter-efficient adaptation, or prompt tuning. This unified workflow enables communication, sensing, localization, and network optimization to share a common representation, substantially improving scalability, transferability, and data efficiency while reducing retraining costs.

\subsection{Motivation and Contributions}
Despite these rapid advances, research on wireless foundation models remains fragmented across different communities and application domains. Existing surveys primarily focus on large AI models for communications, transfer learning, multimodal wireless intelligence, or specific communication applications, without treating WFMs as a unified learning paradigm \cite{wang2026fm, decan2026ai}. Consequently, there remains a lack of a comprehensive survey that systematically integrates architectural design, self-supervised pre-training, adaptation strategies, datasets, benchmark frameworks, evaluation methodologies, deployment considerations, and emerging wireless applications within a single coherent framework. Furthermore, the rapid emergence of representative WFMs since 2024 has fundamentally reshaped the research landscape, highlighting the need for a timely and comprehensive survey dedicated specifically to wireless foundation models.  

Motivated by these developments, this survey presents a unified treatment of wireless foundation models, covering their fundamental principles, architectural design, learning paradigms, adaptation strategies, deployment considerations, evaluation methodologies, and applications across AI-native 6G networks. Beyond summarizing the existing literature, this survey synthesizes the rapidly evolving research landscape, identifies common design principles, compares representative models and learning strategies, discusses emerging research trends, and outlines future directions toward scalable and trustworthy general-purpose wireless intelligence.
The main contributions of this survey are summarized as follows:

\begin{itemize}

\item We establish a unified conceptual framework for wireless foundation models by formalizing their definition, distinguishing them from conventional task-specific deep learning, and describing the transition toward large-scale transferable representation learning for AI-native 6G networks.

\item We develop a comprehensive taxonomy that organizes wireless foundation models from three complementary perspectives, namely model architectures, learning paradigms, and applications, providing a systematic framework for analyzing existing developments and future research directions.

\item We present a comprehensive review and critical comparison of representative wireless foundation models, covering Transformer-based, physics-informed, graph-based, and multimodal architectures together with self-supervised pre-training, parameter-efficient adaptation, prompt-based learning, and transfer learning strategies.

\item We systematically review publicly available datasets, simulation platforms, benchmark methodologies, and evaluation metrics, and discuss their roles in enabling scalable, transferable, and reproducible wireless foundation models.

\item We provide a holistic analysis of emerging applications spanning physical-layer signal processing, integrated sensing and communication, localization, resource management, and network intelligence, highlighting how a shared pre-trained backbone enables general-purpose wireless intelligence across heterogeneous wireless tasks.

\item We identify key research challenges and future opportunities, including data scarcity, domain generalization, multimodal learning, interpretability, efficient edge deployment, continual adaptation, and standardization, and present a research roadmap toward scalable, trustworthy, and AI-native 6G wireless systems.

\end{itemize}

\subsection{Organization}

The remainder of this survey is organized as follows. Section II reviews the background of model-based and learning-based wireless communications together with the fundamental concepts of wireless foundation models. Section III introduces a unified taxonomy of wireless foundation models from the perspectives of architecture, training strategy, and protocol layer. Section IV reviews representative model architectures, while Section V discusses large-scale pre-training strategies and adaptation techniques. Section VI summarizes the major application domains of wireless foundation models. Section VII presents publicly available datasets, benchmark platforms, and evaluation methodologies. Section VIII discusses the major open research challenges and future research directions, followed by a roadmap toward AI-native 6G systems in Section IX. Finally, Section X concludes the survey.

\section{Evolution Toward Wireless Foundation Models}
WFMs represent the latest stage in the evolution of artificial intelligence for wireless communications. Their emergence has been driven by the limitations of both conventional model-based signal processing and task-specific deep learning to address the increasing complexity of AI-native 6G networks \cite{rahmani2026hybrid}. Understanding this evolution provides the necessary context to appreciate the design philosophy of WFMs and the motivation behind the large-scale transferable representation learning.

This section first reviews the foundations of model-based wireless communications and the transition toward deep learning. It then discusses how the success of foundation models in natural language processing (NLP) and computer vision (CV) has inspired a new generation of wireless learning systems capable of supporting multiple communication tasks through a shared pre-trained backbone.

\subsection{Model-Based Wireless Communications}

Wireless communication systems have traditionally relied on analytical models, estimation theory, optimization, and statistical signal processing to solve fundamental problems such as channel estimation, signal detection, synchronization, beamforming, equalization, and resource allocation \cite{guo2022private}. Classical approaches, including least-squares (LS), minimum mean square error (MMSE), maximum likelihood (ML), and convex optimization, provide mathematically interpretable solutions with well-understood theoretical guarantees and have formed the cornerstone of successive wireless generations \cite{Tse2005Fundamentals,Goldsmith2005Wireless}.

The effectiveness of these techniques depends on accurate mathematical descriptions of wireless propagation and network behavior. For example, channel estimation algorithms assume specific statistical channel models, while beamforming and resource allocation require reliable CSI and tractable optimization formulations \cite{albataineh2026beam}. Under these assumptions, model-driven approaches achieve predictable performance and can often be analyzed rigorously using estimation theory, information theory, and convex optimization \cite{Kay1993Estimation,Boyd2004Convex}.

However, the transition toward AI-native 6G networks has significantly increased the complexity of wireless systems \cite{javaid2026post}. Emerging technologies, including massive MIMO, millimeter-wave and terahertz communications, ISAC, ultra-dense heterogeneous deployments, and highly dynamic propagation environments, introduce nonlinear interactions and uncertainties that are increasingly difficult to capture using analytical models alone\cite{qaisar2026role}. Consequently, deriving accurate mathematical models and designing optimal signal processing algorithms have become considerably more challenging, motivating a gradual transition toward data-driven learning approaches capable of extracting knowledge directly from wireless measurements.

\subsection{Deep Learning for Wireless Communications}

Deep learning has transformed wireless communications by enabling data-driven models to learn complex nonlinear relationships directly from wireless observations, complementing conventional approaches that rely on analytical channel and signal models \cite{Goodfellow2016DeepLearning}. Its effectiveness has been demonstrated across a broad range of physical-layer tasks, including channel estimation, signal detection, beamforming, localization, modulation recognition, and resource allocation \cite{Dai2020DLWireless}. For example, deep neural networks have been successfully applied to joint channel estimation and signal detection in OFDM systems, illustrating their ability to learn communication functions directly from received signals \cite{Ye2018DLofdm}. Such learning capabilities are particularly relevant as wireless systems evolve toward increasingly high-dimensional configurations involving technologies such as massive MIMO and millimeter-wave communications \cite{Boccardi2014Five5G}.

Despite these advances, current learning-based wireless systems remain predominantly task-specific. Individual models are typically trained for a single communication task under particular channel conditions and deployment assumptions. Consequently, their performance often degrades under changes in propagation environments, antenna configurations, user mobility, hardware impairments, or other distribution shifts \cite{Liu2021OODWireless}. Recovering performance usually requires collecting additional labeled data and retraining or fine-tuning the model, limiting scalability in practical deployments.

Another major limitation is the dependence on large labeled datasets. Accurate supervision often requires pilot-assisted measurements, extensive simulations, or costly measurement campaigns, making dataset construction both expensive and time-consuming \cite{Nguyen2021TransferLearningWireless}. In addition, modern deep neural networks typically require substantial computational resources, memory, and training time, posing significant challenges for latency-sensitive and resource-constrained wireless devices. Their black-box nature also limits interpretability and complicates verification in safety-critical wireless applications \cite{Doha2025DLWirelessReceivers}.

Collectively, these limitations highlight that task-specific deep learning alone is insufficient to support the scalability, adaptability, and generalization required by AI-native 6G networks. Rather than repeatedly training independent models for every communication task, a more scalable paradigm is needed one that learns reusable wireless representations capable of transferring knowledge across heterogeneous tasks and deployment scenarios. This requirement has motivated the emergence of wireless foundation models.
\section{Definitions and Taxonomy of Wireless Foundation Models}
\begin{figure*}[t]
\centering
\includegraphics[width=0.9\textwidth]{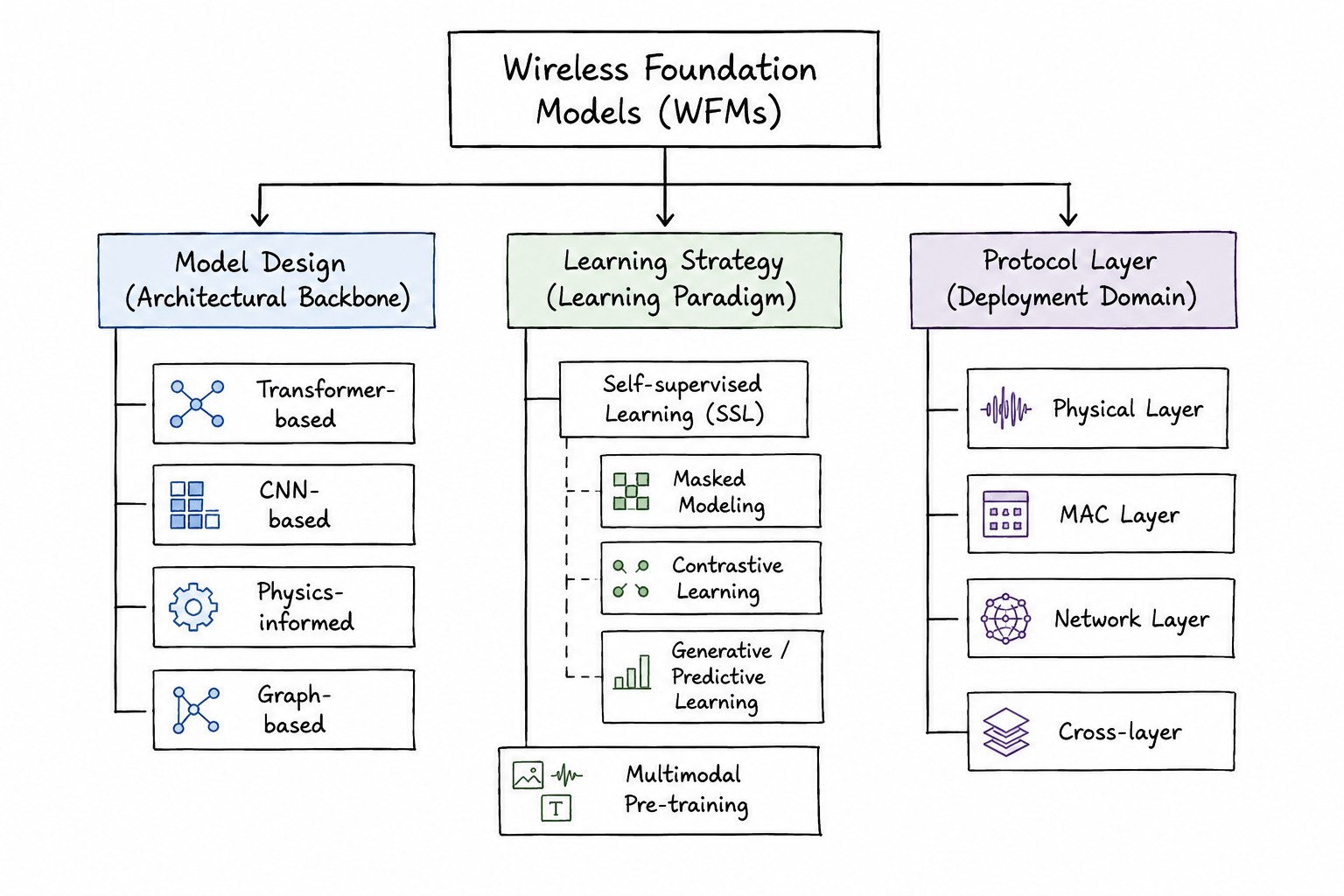}
\caption{Taxonomy of wireless foundation models based on three complementary dimensions: (i) model architecture; (ii) pre-training strategy; and (iii) deployment, including physical-layer, MAC-layer, network-layer, and cross-layer optimization tasks.}
\label{fig:wfm_taxonomy}
\end{figure*}

WFMs represent a new generation of learning-based wireless intelligence that extends beyond conventional task-specific deep learning. Instead of developing independent models for individual communication tasks, a wireless foundation model is pre-trained on large-scale heterogeneous wireless data to learn transferable representations that can be efficiently adapted to multiple downstream applications\cite{pan2025large}. Owing to the diversity of existing model architectures, pre-training strategies, and application domains, a systematic taxonomy is essential for understanding the rapidly evolving research landscape.
As illustrated in Fig.~\ref{fig:wfm_taxonomy}, wireless foundation models can be classified from three complementary perspectives: \emph{(i)} model architecture, which determines how wireless representations are learned; \emph{(ii)} pre-training strategy, which defines how transferable knowledge is acquired from large-scale wireless datasets; and \emph{(iii)} deployment layer, which categorizes the downstream communication and networking tasks supported by the pre-trained model.

The proposed taxonomy is organized around these three dimensions because they collectively describe the complete lifecycle of a wireless foundation model. The architectural dimension specifies the underlying neural backbone responsible for representation learning, the pre-training dimension characterizes how transferable knowledge is acquired from large-scale wireless data, and the application dimension reflects how the learned representations are adapted to downstream wireless communication and networking tasks. Other possible classifications, such as data modality or deployment platform, typically represent specific aspects of these three fundamental dimensions rather than independent design categories. Therefore, this taxonomy provides a unified and systematic framework for analyzing, comparing, and organizing existing wireless foundation models while accommodating future developments in AI-native 6G systems.

\subsection{Definition of Wireless Foundation Models}

WFMs extend conventional task-specific learning by introducing a unified pre-trained backbone that can be adapted to diverse wireless applications. Unlike traditional machine learning and deep learning models, which are typically optimized for a single communication task or deployment scenario, WFMs learn transferable representations from large-scale heterogeneous wireless data and reuse the acquired knowledge across multiple downstream tasks with limited task-specific supervision \cite{Buffelli2025CommFM}. This paradigm enables a single model to support communication, sensing, localization, and network intelligence within a unified learning framework, thereby improving scalability, data efficiency, and cross-task generalization.

The operation of a WFM follows a two-stage learning paradigm consisting of large-scale pre-training and downstream adaptation. During the pre-training stage, heterogeneous wireless measurements, including CSI, IQ samples, CIRs, RF signals, spectrograms, sensing measurements, and network observations, are used to learn a shared wireless representation through self-supervised or unsupervised objectives such as masked reconstruction, contrastive learning, or generative modeling \cite{Yang2025WirelessGPT,Aboulfotouh2025WavesFM,Alikhani2025LWM}. Rather than learning task-specific features, the objective is to capture common spatial, temporal, spectral, and propagation characteristics that are transferable across different wireless environments.

The learned representation is subsequently adapted to downstream applications using lightweight techniques, including full fine-tuning, parameter-efficient adaptation, prompt tuning, or task-specific prediction heads. Representative applications include channel estimation, channel prediction, beamforming, MIMO detection, CSI feedback, localization, integrated sensing, wireless signal classification, and resource allocation \cite{Zheng2025MUSEFM,Cheraghinia2025WTRLocalization}. Compared with independently trained deep learning models, this shared-backbone paradigm substantially reduces labeled data requirements, retraining costs, and deployment complexity while improving robustness across heterogeneous wireless scenarios.

Fig.~\ref{fig:wfm_working_mechanism} summarizes the fundamental workflow of a wireless foundation model. Large-scale self-supervised pre-training first learns a shared representation from heterogeneous wireless data, after which the pre-trained backbone is efficiently adapted to diverse downstream communication tasks through lightweight adaptation mechanisms. This separation of representation learning from task-specific optimization constitutes the defining characteristic of WFMs and provides the conceptual basis for the taxonomy presented in the remainder of this section.

\subsection{Taxonomy by Architecture}

The architectural backbone of a wireless foundation model largely determines its representation learning capability, scalability, computational efficiency, and adaptation performance. As illustrated in Fig.~\ref{fig:wfm_taxonomy}, current WFMs can be broadly categorized into four architectural families: Transformer-based, convolutional neural network (CNN)-based, physics-informed, and graph-based models. Rather than representing competing solutions, these architectures offer complementary design philosophies for learning generalized wireless representations across different communication scenarios.
Transformer-based architectures have emerged as the dominant backbone for wireless foundation models owing to their ability to model long-range spatial, temporal, and frequency-domain dependencies through self-attention mechanisms\cite{mao2026biolamr}. Their scalability and strong representation learning capability make them particularly well suited for large-scale self-supervised pre-training using heterogeneous wireless measurements, CSI, IQ samples, CIRs, spectrograms, and RF signals. Consequently, most recent WFMs, including WirelessGPT, LWM, and WavesFM, adopt Transformer-based architectures as their primary learning backbone \cite{Vaswani2017Attention,Dosovitskiy2021ViT}.

CNN-based architectures remain attractive for applications where computational efficiency and low-latency inference are critical. By exploiting local spatial and temporal correlations, CNNs provide lightweight yet effective feature extraction for tasks such as channel estimation, modulation recognition, spectrum sensing, and wireless signal classification. Although they generally exhibit lower representation capacity than Transformers, their efficiency makes them well suited for edge deployment and resource-constrained wireless devices \cite{LeCun2015Deep,Goodfellow2016DeepLearning}.
Physics-informed architectures combine data-driven learning with communication-domain knowledge by embedding analytical models, optimization algorithms, or signal processing principles into trainable neural networks. Representative approaches, such as deep unfolding, preserve the interpretability of conventional communication algorithms while improving robustness, sample efficiency, and generalization. These models are particularly attractive for wireless tasks where reliable physical models are available but require adaptive learning capabilities \cite{He2018ModelDriven,Hershey2014DeepUnfolding}.

Graph-based architectures naturally represent wireless networks as graphs, where nodes correspond to users, base stations, access points, or network entities, and edges describe communication, interference, or connectivity relationships. By explicitly modeling network topology, graph neural networks effectively capture spatial interactions that are difficult to represent using conventional neural architectures, making them particularly suitable for resource allocation, routing, user association, interference management, and network optimization in large-scale AI-native 6G systems \cite{Wu2021Graph,Liu2023GraphWireless}.
Fig.~\ref{fig:wfm_taxonomy} highlights that these architectural families differ primarily in how they encode wireless information rather than in their overall learning objective. Transformer models emphasize global contextual modeling, CNNs focus on local feature extraction, physics-informed models integrate communication-domain knowledge, whereas graph-based models explicitly exploit network topology. In practice, emerging WFMs increasingly combine multiple architectural paradigms to balance representation quality, computational efficiency, interpretability, and scalability, suggesting that future wireless foundation models are likely to evolve toward hybrid architectures rather than relying on a single neural backbone.
\subsection{Taxonomy by Learning Paradigm}

Besides architectural design, wireless foundation models can also be categorized according to their learning paradigm, which determines how transferable wireless representations are acquired from large-scale heterogeneous data. Unlike conventional supervised learning, where models are optimized for a single task using manually annotated datasets, WFMs primarily rely on self-supervised pre-training to exploit the abundance of unlabeled wireless measurements. As illustrated in Fig.~\ref{fig:wfm_taxonomy}, current learning paradigms can be broadly grouped into self-supervised learning, contrastive representation learning, masked modeling, and multimodal pre-training \cite{ericsson2022self,Zhang2026MultiModalSurvey}.

Self-supervised learning (SSL) forms the foundation of most contemporary WFMs. Instead of relying on manually generated labels, SSL derives supervisory signals directly from the input data through carefully designed pretext tasks. This learning paradigm is particularly attractive for wireless communications because large volumes of CSI, IQ samples, RF signals, spectrograms, and channel impulse responses are continuously generated during network operation, whereas obtaining accurate labels often requires costly measurements, simulations, or annotation. Recent WFMs illustrate the effectiveness of this approach across different wireless modalities. WirelessGPT \cite{Yang2025WirelessGPT} and LWM \cite{Alikhani2025LWM} exploit self-supervised pre-training to learn transferable representations from heterogeneous wireless observations, while WavesFM \cite{Aboulfotouh2025WavesFM} extends representation learning across diverse wireless signals and tasks. Similarly, IQFM \cite{Mashaal2025IQFM} focuses on reusable representations from IQ data, whereas CSI2Vec \cite{Palhares2025CSI2Vec} learns transferable representations from CSI. Collectively, these models demonstrate that self-supervised pre-training can reduce dependence on task-specific labels while enabling learned representations to be reused across multiple downstream communication tasks.

Within the SSL framework, different pre-training objectives encourage models to capture complementary characteristics of wireless signals. Contrastive learning, in particular, learns discriminative representations by increasing the similarity between related wireless observations while separating unrelated samples in the latent space. Positive pairs are typically constructed from different augmentations of the same signal or measurements obtained under similar propagation conditions, whereas negative pairs correspond to unrelated channel or signal realizations. Such objectives can improve representation robustness to signal and channel variations and facilitate transfer across downstream tasks. ContraWiMAE \cite{Guler2025ContraWiMAE}, for example, combines contrastive principles with masked representation learning, while CSI2Vec \cite{Palhares2025CSI2Vec} applies representation learning specifically to CSI. Related approaches extend this principle to other wireless modalities, with IQFM targeting IQ signals \cite{Mashaal2025IQFM} and CSI-CLIP exploiting contrastive alignment for CSI representations \cite{Jiang2025CSICLIP}. These developments illustrate how contrastive objectives can be tailored to different wireless data modalities while retaining a common goal of learning transferable and discriminative representations.

Masked modeling has recently emerged as an effective self-supervised objective for WFMs. Inspired by masked language modeling and masked image modeling, portions of the wireless input are intentionally hidden and the model is trained to reconstruct the missing information. This approach is well suited to wireless data because signals often exhibit strong spatial, temporal, and frequency-domain correlations, allowing models to learn propagation and signal structure by reconstructing masked CSI matrices, IQ sequences, OFDM resource grids, or spectrogram patches without requiring explicit labels. WavesFM, for example, employs reconstruction-oriented pre-training to learn generalizable representations from wireless signals \cite{Aboulfotouh2025WavesFM}. Scalable masked channel modeling further exploits the inherent structure of channel observations to support transferable channel representations \cite{Guo2026MaskedChannelModel}, while WiFo applies masked pre-training to learn reusable wireless features for channel-related downstream tasks \cite{Liu2025WiFo}. Collectively, these approaches demonstrate the potential of masked modeling for applications such as channel estimation, channel prediction, and CSI feedback.

Beyond single-modality learning, multimodal pre-training extends representation learning by jointly modeling heterogeneous wireless and contextual information. This capability is particularly relevant to future AI-native 6G networks, where communication, sensing, localization, and network management are expected to become increasingly integrated \cite{Zhang2026MultiModalSurvey}. Such systems may need to jointly exploit wireless modalities such as CSI, IQ samples, and CIRs together with radar measurements, LiDAR, RGB images, GPS information, traffic statistics, and environmental context. Multimodal WFM explores the integration of heterogeneous wireless information within a shared representation space \cite{Aboulfotouh2025MultimodalWFM}, while MuSE-FM extends this principle toward multimodal sensing and environmental representations \cite{Zheng2025MUSEFM}. CSI-CLIP further demonstrates contrastive alignment between CSI and complementary modalities, enabling semantically aligned representations across heterogeneous observations \cite{Jiang2025CSICLIP}. These developments suggest that multimodal pre-training can provide richer representations by exploiting complementary information that is unavailable to models trained on individual modalities alone.
\begin{figure*}[t]
    \centering
    \includegraphics[width=0.98\textwidth]{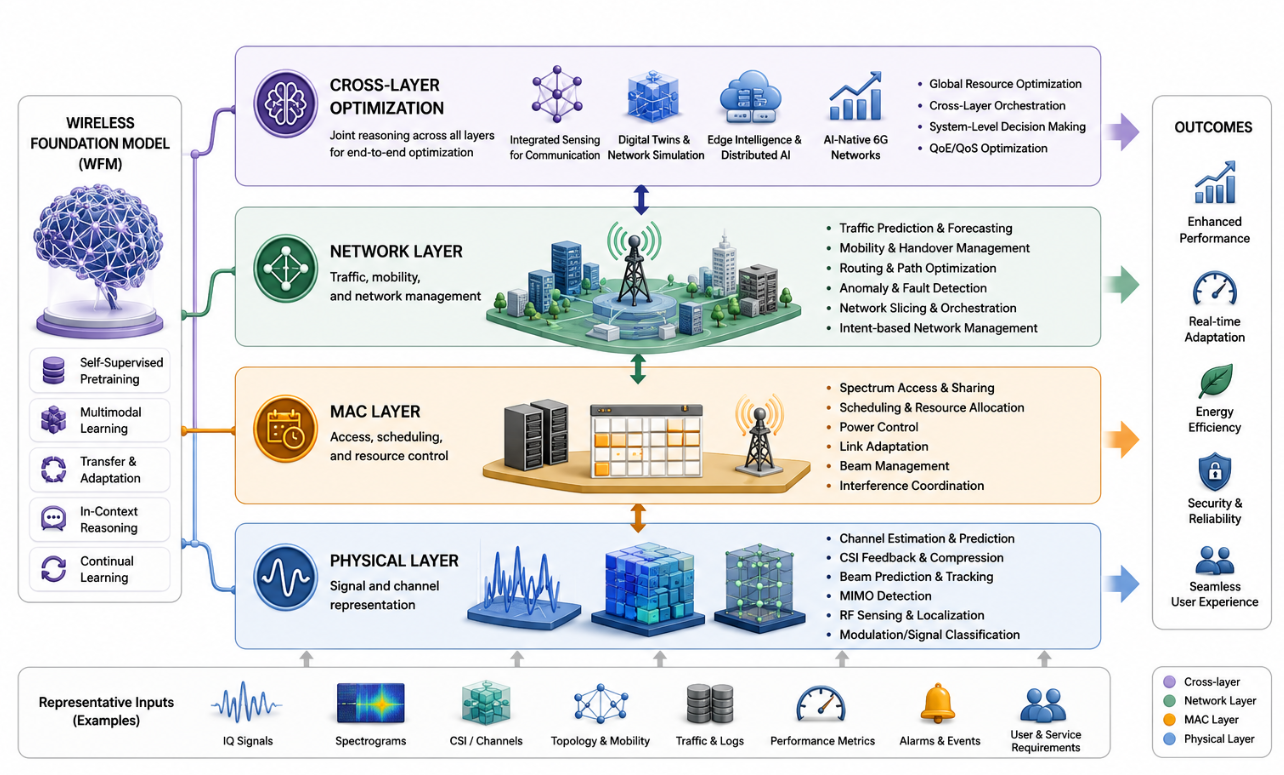}
    \caption{Protocol-layer taxonomy of wireless foundation models. A shared pre-trained backbone is adapted across the wireless protocol stack, enabling communication, resource management, network intelligence, and cross-layer optimization through transferable wireless representations.}
    \label{fig:wfm_application_layers}
\end{figure*}
The learning paradigms shown in Fig.~\ref{fig:wfm_taxonomy} should therefore be viewed as complementary rather than mutually exclusive. Contemporary WFMs frequently combine multiple objectives during pre-training, e.g, integrating masked reconstruction with contrastive representation learning or multimodal alignment to improve robustness, transferability, and generalization. Consequently, the evolution of WFMs is moving toward unified pre-training frameworks that simultaneously exploit multiple learning objectives instead of relying on a single self-supervised strategy.

\subsection{Taxonomy by Deployment}
Besides architectural design and learning paradigm, wireless foundation models can also be categorized according to the layer of the wireless protocol stack in which they are deployed. This perspective is particularly important because each protocol layer operates on different data sources, optimization objectives, latency requirements, and decision timescales. As illustrated in Fig.~\ref{fig:wfm_application_layers}, a shared pre-trained backbone can support applications spanning the entire wireless stack, from physical-layer signal processing to network-wide intelligence and cross-layer optimization \cite{Liang2026FMWireless}.
Fig.~\ref{fig:wfm_application_layers} highlights that WFMs enable a common representation to be reused across protocol layers rather than developing independent AI models for individual networking functions. This shared representation allows knowledge acquired from one wireless task to improve performance in related applications, thereby reducing retraining effort while improving scalability and transferability throughout AI-native 6G systems.

At the physical layer, WFMs primarily support signal processing tasks that operate directly on wireless measurements such as CSI, IQ samples, CIRs, and RF signals. Representative applications include channel estimation, channel prediction, CSI feedback, MIMO detection, beam prediction, RF signal classification, localization, and integrated sensing. Since these tasks exhibit strong spatial, temporal, and frequency-domain correlations, they benefit significantly from large-scale pre-training and transferable representations. Consequently, the physical layer has become the most mature application domain for WFMs, with representative systems including WirelessGPT, LWM, WavesFM, and WiFo demonstrating substantial improvements in transferability and data efficiency \cite{fontaine2024towards, abo2026enhancing}.

Moving beyond signal processing, the MAC layer focuses on intelligent radio resource management. WFMs can learn network-level representations that support adaptive spectrum access, scheduling, beam management, interference coordination, power allocation, and link adaptation. Recent LLM-based approaches, for example, have explored intelligent resource allocation and radio-access optimization using contextual network information \cite{Lee2026LLMResourceAllocation,Noh2025LLMRAO}. Other studies investigate task prediction and adaptation through shared wireless representations \cite{Sheng2025WFMTaskPrediction}, while LLM-assisted MAC frameworks extend this capability to MAC-layer decision making and coordination \cite{Tan2025LLM4MAC}. These developments indicate a shift from independently optimized functions toward shared models that can incorporate network state, user behavior, and service requirements to support more adaptive resource management in dynamic wireless environments.

At the network layer, WFMs operate on broader contextual information, including traffic statistics, mobility patterns, topology information, network telemetry, service requirements, and quality-of-service indicators. Such information can support functions ranging from traffic prediction and mobility management to routing, anomaly detection, network slicing, and intent-driven networking. Digital twin-assisted task offloading, for instance, illustrates how learned network representations and contextual information can facilitate adaptive decisions across cloud, edge, and hybrid deployments \cite{alshareeda2026task}. More broadly, large AI models are increasingly being investigated as a foundation for intelligent wireless network operation \cite{Huang2026LargeAIWireless}, including network management and orchestration \cite{Wei2025LLMNetworkManagement}. Emerging frameworks further extend this direction toward greater wireless autonomy \cite{Liang2026LLMWirelessAutonomy} and world-model-based representations of telecommunication environments \cite{Zou2026TelecomWorldModels}. Together with task-oriented WFMs \cite{Sheng2025WFMTaskPrediction}, these developments point toward increasingly self-managing and self-optimizing AI-native 6G networks.

Although protocol layers are conventionally designed as distinct functional entities, their decisions are inherently coupled. Physical-layer channel conditions influence MAC-layer scheduling and resource allocation, while traffic dynamics and service requirements at higher layers affect radio resource allocation, energy management, and communication reliability. This coupling motivates cross-layer WFMs that jointly represent wireless signals, network state, environmental context, and service objectives. Such cross-layer intelligence is particularly relevant to ISAC, where sensing, communication, and resource-management decisions are closely interconnected \cite{qu2024isac,Wymeersch2025CrossLayerISAC}. Multimodal WFMs provide another pathway by integrating heterogeneous wireless and contextual information within shared representations \cite{Aboulfotouh2025MultimodalWFM,Zheng2025MUSEFM}, while emerging models are extending this concept toward integrated communication and sensing \cite{Liu2026WiFoMiSAC}. Similar cross-layer principles are also relevant to semantic communications, digital twins, edge intelligence, and autonomous network management, where decisions increasingly span multiple functional layers and timescales \cite{Zou2026TelecomWorldModels}.

\section{Architectures for Wireless Foundation Models}

The architectural backbone of a wireless foundation model largely determines its representation learning capability, computational efficiency, scalability, and adaptation performance. Unlike conventional deep learning models that are designed for specific communication tasks, WFM architectures aim to learn transferable wireless representations that can be efficiently reused across diverse downstream applications\cite{shi2026unified}. Although several neural architectures have been explored, Transformer-based models have emerged as the dominant backbone owing to their superior ability to capture long-range dependencies and scale to large self-supervised pre-training datasets. Meanwhile, physics-informed, graph-based, and lightweight architectures complement Transformers by incorporating communication-domain knowledge, network topology, or computational efficiency\cite{liu2025efficient}. This section reviews these architectural paradigms and discusses their suitability for AI-native 6G systems.

\subsection{Transformer-Based Architectures}

Transformer-based architectures have become the primary backbone of wireless foundation models because they combine scalable representation learning with excellent transferability across heterogeneous wireless tasks. Their self-attention mechanism enables the joint modeling of spatial, temporal, and frequency-domain correlations that naturally exist in wireless measurements, making them particularly suitable for learning generalized representations from large-scale heterogeneous datasets \cite{ jing2026signal}.

Unlike conventional task-specific neural networks, Transformer-based WFMs separate large-scale representation learning from downstream task adaptation. During self-supervised pre-training, the model learns reusable wireless representations from heterogeneous measurements, including CSI, IQ samples, CIRs, spectrograms, and RF signals\cite{cheraghinia2025foundation}. These representations are subsequently adapted to diverse communication, sensing, and networking applications through lightweight techniques such as full fine-tuning, Low-Rank Adaptation (LoRA), adapters, or prompt tuning, substantially reducing labeled data requirements and retraining costs \cite{abdullah2025evolution}.
The flexibility of the Transformer architecture allows it to accommodate multiple wireless data modalities. Sequential representations, such as IQ samples and CSI time series, are naturally processed using sequence Transformers, whereas image-like representations, including CSI matrices, spectrograms, beamspace maps, and orthogonal frequency-division multiplexing (OFDM) resource grids, are effectively modeled using Vision Transformers (ViTs) \cite{Dosovitskiy2021ViT}. This unified representation framework enables the direct application of self-supervised objectives such as masked modeling and contrastive learning, allowing WFMs to exploit large volumes of unlabeled wireless data while learning representations that generalize across multiple propagation environments and communication tasks.

Several representative models illustrate the potential of this paradigm. WirelessGPT employs large-scale self-supervised pre-training to learn general-purpose wireless representations that can be adapted to different downstream tasks \cite{Yang2025WirelessGPT}. LWM similarly explores transferable representation learning for channel modeling and broader communication intelligence \cite{Alikhani2025LWM}. WavesFM extends this direction by combining a Vision Transformer backbone with masked pre-training and LoRA-based adaptation to support communication, sensing, and localization tasks within a shared framework \cite{Aboulfotouh2025WavesFM}. Collectively, these studies highlight the potential of Transformer-based WFMs to improve representation reuse, parameter-efficient adaptation, and transferability across wireless tasks compared with independently trained task-specific models.

Despite these advantages, several challenges remain. The quadratic computational and memory complexity of global self-attention can limit scalability when processing long CSI sequences, high-dimensional channel observations, or large wireless resource grids. Moreover, standard Transformer architectures do not inherently encode communication-domain characteristics such as sparse multipath propagation, delay-Doppler structure, channel reciprocity, and antenna geometry \cite{balaji2025spatio}. Efficient attention mechanisms have therefore been investigated to reduce the computational burden associated with conventional self-attention \cite{Tay2023Efficient}. Within wireless systems, emerging architectures increasingly incorporate scalable and domain-aware designs to better capture the structure of wireless observations, as exemplified by AirFM \cite{Bian2026AirFM}. Lightweight Transformer variants provide another direction for reducing inference and deployment overhead in resource-constrained wireless environments \cite{Cheraghinia2025Lightweight}. These developments suggest that future Transformer-based WFMs will increasingly combine scalable attention with wireless-domain inductive biases to support efficient representation learning for AI-native 6G systems.
\begin{figure*}[t]
    \centering
    \includegraphics[width=\textwidth]{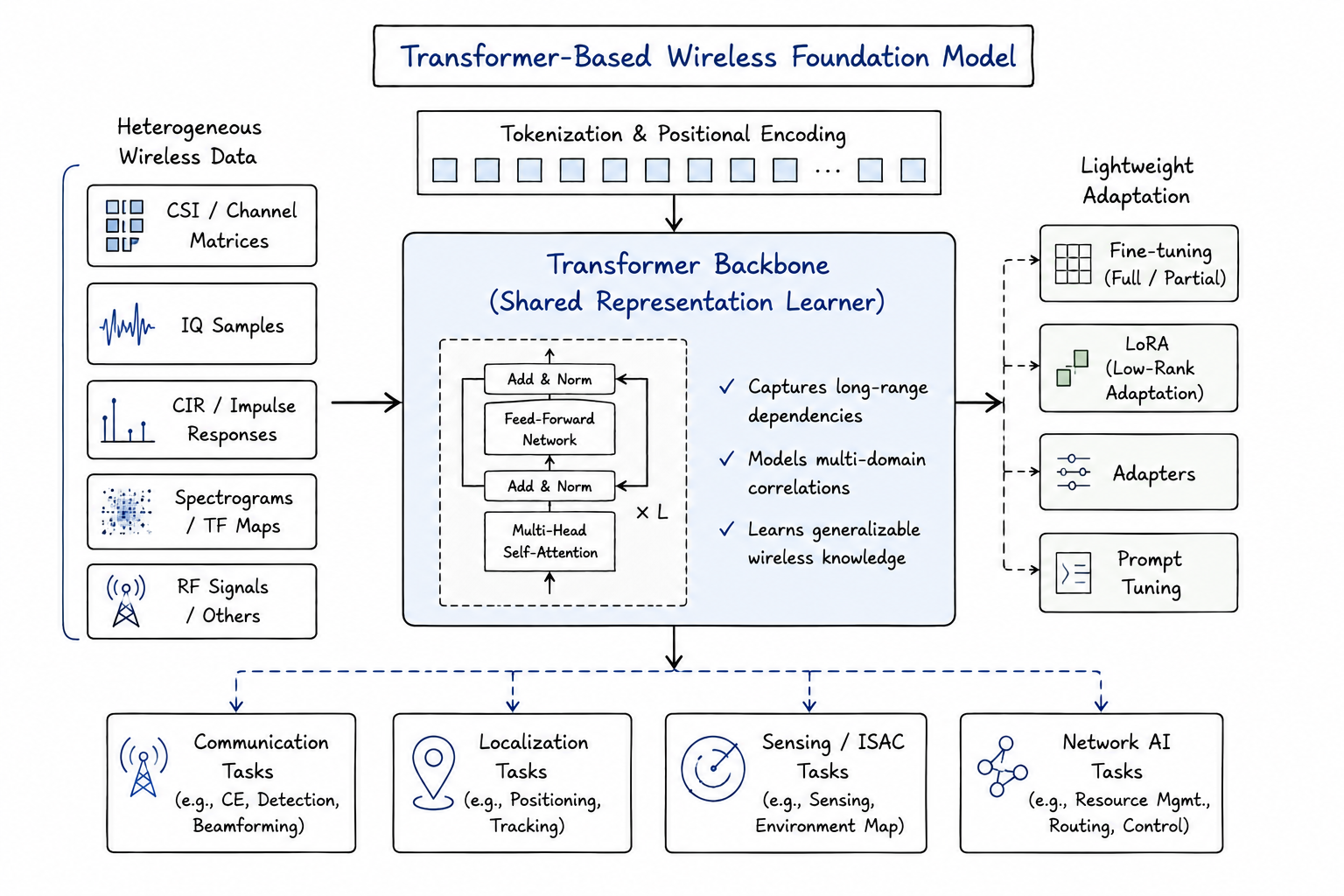}
    \caption{Conceptual architecture of a Transformer-based wireless foundation model.}
    \label{fig:transformer_wfm}
\end{figure*}
Fig.~\ref{fig:transformer_wfm} summarizes the learning paradigm of a Transformer-based wireless foundation model. Instead of functioning as a task-specific predictor, the Transformer serves as a reusable representation learner that acquires generalized wireless knowledge during large-scale self-supervised pre-training. The resulting backbone is subsequently specialized to downstream communication, sensing, and localization tasks through lightweight adaptation mechanisms, enabling a single pre-trained model to support diverse wireless applications while preserving the knowledge acquired during pre-training. This decoupling of representation learning from task-specific optimization constitutes the defining characteristic of Transformer-based wireless foundation models.

\subsection{Physics-Informed and Model-Driven Architectures}

Although Transformer-based architectures provide powerful representation learning capabilities, purely data-driven models often require large training datasets and may struggle to generalize under unseen propagation environments. Wireless communication systems, however, are governed by well-established physical principles, including channel propagation, estimation theory, signal detection, antenna array processing, and optimization. Physics-informed and model-driven architectures seek to bridge these complementary strengths by integrating communication-domain knowledge into modern foundation models, thereby improving data efficiency, interpretability, robustness, and generalization while preserving the scalability of large-scale representation learning \cite{He2018ModelDriven,Shlezinger2023ModelDriven}.

The central idea is to embed analytical models and physical constraints directly into the learning process rather than relying solely on statistical correlations. Early model-driven approaches achieved this through deep unfolding, where iterative communication algorithms were transformed into trainable neural network layers. Representative examples include unfolded approximate message passing (AMP), projected gradient descent (PGD), and weighted minimum mean-square error (WMMSE), which preserve the mathematical structure of conventional optimization algorithms while allowing trainable parameters to be learned from data \cite{Hershey2014DeepUnfolding,He2018ModelDriven}. These hybrid architectures demonstrated that incorporating communication-domain knowledge significantly improves convergence, robustness, and sample efficiency compared with purely data-driven models.

Recent wireless foundation models extend this philosophy beyond individual communication tasks by incorporating physical priors directly into large-scale self-supervised pre-training. Rather than learning generic statistical representations, these models exploit communication-specific characteristics such as channel reciprocity, sparse multipath propagation, delay-Doppler structure, antenna geometry, and optimization constraints to guide representation learning toward physically meaningful solutions that generalize across heterogeneous deployment scenarios \cite{Bian2026AirFM,Zhang2026Adaptive3DRoPE}.
Representative systems illustrate this evolution. AirFM-DDA learns wireless representations in the delay-Doppler-angle domain to better capture propagation characteristics, while Adaptive 3D-RoPE introduces physics-aware positional encoding that preserves spatial and temporal relationships within wireless measurements. These studies demonstrate that integrating communication-domain knowledge into Transformer-based foundation models substantially improves representation quality and robustness without sacrificing scalability \cite{Bian2026AirFM,Zhang2026Adaptive3DRoPE}.



\subsection{Multimodal Architectures}

Future AI-native 6G networks are expected to simultaneously support communication, sensing, localization, and network intelligence, requiring wireless foundation models to process information from multiple heterogeneous sources rather than a single wireless modality. Consequently, multimodal architectures have emerged as an important extension of Transformer-based foundation models by jointly learning representations from communication signals and contextual information. Compared with unimodal architectures, multimodal models provide a richer understanding of the wireless environment, enabling improved generalization, robustness, and task transfer across diverse wireless applications \cite{10599304, 11495157}.
Modern multimodal wireless foundation models integrate various input modalities, including CSI, IQ samples, CIR, spectrograms, RF signals, localization information, and environmental observations. These complementary modalities capture different aspects of the wireless environment, allowing the model to jointly exploit spatial, temporal, spectral, and contextual information that cannot be fully represented by a single data source \cite{11315904}.
Representative architectures demonstrate different approaches to multimodal representation learning. WavesFM adopts a shared Vision Transformer (ViT) backbone capable of processing image-like wireless representations, such as CSI matrices, spectrograms, and OFDM resource grids, while employing LoRA for parameter-efficient task adaptation \cite{Aboulfotouh2025WavesFM}. WirelessGPT, in contrast, focuses on learning generalized channel representations through large-scale self-supervised pre-training, enabling efficient transfer across multiple communication and sensing tasks \cite{11036155}. More recent multimodal wireless foundation models further extend this paradigm by incorporating complementary information from radar, LiDAR, cameras, GPS, network telemetry, and environmental measurements, thereby improving situational awareness and enabling joint communication, sensing, and localization \cite{Jiang2025ComLAMSurvey,Zhang2026MultiModalSurvey}.

Architecturally, multimodal foundation models differ from conventional wireless models because they require dedicated modality encoders together with fusion mechanisms that align heterogeneous feature spaces before learning a shared representation. Early fusion, late fusion, cross-attention, and multimodal Transformer encoders are among the most commonly adopted strategies for integrating complementary information while preserving modality-specific characteristics. The resulting shared representation enables efficient adaptation to multiple downstream tasks without requiring separate models for each sensing modality.

Despite their considerable potential, multimodal architectures introduce additional challenges, including modality alignment, synchronization, missing or incomplete observations, increased computational complexity, and limited availability of large-scale multimodal datasets. Addressing these issues will require scalable multimodal pre-training strategies, efficient cross-modal representation learning, and standardized benchmark datasets. These developments are expected to play a key role in enabling robust and context-aware wireless foundation models for future AI-native 6G systems.

\subsection{Scalability Considerations}

Scalability is a fundamental requirement for practical wireless foundation models because future AI-native 6G networks must support diverse communication tasks, heterogeneous wireless environments, and resource-constrained deployment platforms. Unlike conventional deep learning models designed for a single communication scenario, wireless foundation models are expected to operate across different channel configurations, frequency bands, antenna arrays, sampling rates, and sensing modalities while maintaining efficient training and inference. Consequently, scalability should be evaluated not only in terms of model accuracy, but also with respect to parameter count, memory consumption, computational complexity, adaptation cost, and inference latency, all of which directly influence practical deployment at cloud servers, base stations, edge nodes, and user equipment \cite{Cheng2026LWFM,Huang2026LargeAIWireless,Du2024DistributedFM6G}.
Model size and memory consumption remain among the most important scalability considerations. Larger backbone models generally provide stronger representation learning and improved transferability across wireless tasks; however, they also increase storage requirements, communication overhead, and fine-tuning costs. 

To address these challenges, recent architectures increasingly combine shared pre-trained backbones with parameter-efficient adaptation techniques, allowing multiple downstream tasks to reuse a common model without full retraining. LoRA, for example, updates only low-rank parameter components and substantially reduces the number of trainable parameters during adaptation \cite{Hu2021LoRA}. This strategy has been incorporated into WFMs such as WavesFM to enable efficient adaptation across different wireless tasks \cite{Aboulfotouh2025WavesFM}, while related multimodal frameworks, including MuSE-FM, further explore efficient adaptation of shared representations across heterogeneous inputs \cite{Zheng2025MUSEFM}. Beyond adaptation, model compression provides another pathway toward practical deployment. Pruning, quantization, and other lightweight model-design techniques can reduce memory, computation, and inference overhead, as demonstrated by compact wireless foundation models such as TinyWiFo \cite{Zhang2025TinyWiFo} and TinyFWFM \cite{Hallaq2025TinyFWFM}. These techniques are particularly important for deploying WFMs on resource-constrained edge and IoT devices, where computational and energy budgets are limited \cite{Zhang2026MultiModalSurvey}.

Training scalability presents another important challenge because WFMs require large and heterogeneous datasets together with substantial computational resources during pre-training. Variations across propagation environments, hardware platforms, frequency bands, and communication standards further complicate the learning of representations that remain transferable across deployment conditions. One emerging approach is to incorporate communication-domain structure directly into the learning process. Masked channel modeling exploits correlations within wireless channel observations to improve representation learning \cite{Guo2026MaskedChannelModel}, while delay-Doppler-aware architectures incorporate the underlying structure of time-varying wireless channels \cite{Bian2026AirFMDDA}. Physics-aware positional encoding, including adaptive three-dimensional positional representations, provides another mechanism for embedding spatial and propagation structure into the model architecture \cite{Zhang2026Adaptive3DRoPE}. Such domain-aware designs can reduce the burden on purely data-driven learning by introducing inductive biases that better reflect the physical structure of wireless environments.

Inference efficiency is equally critical because many wireless decisions must be completed within channel coherence times or stringent scheduling intervals. Although standard Transformer architectures provide powerful representation-learning capabilities, the quadratic computational and memory complexity of global self-attention can become prohibitive for long wireless sequences and high-dimensional observations \cite{Tay2023EfficientTransformers}. Consequently, recent WFMs increasingly explore architectures that reduce inference overhead while preserving the ability to capture long-range dependencies. WiMamba, for example, employs state-space modeling as an alternative to conventional attention for efficient processing of wireless observations \cite{Raviv2026WiMamba}. ComHymba adopts a hybrid architecture that combines complementary sequence-modeling mechanisms to balance representation capability and computational efficiency \cite{Yang2026ComHymba}. AirFM further incorporates structured and efficient processing tailored to wireless channel representations, including delay-Doppler-aware modeling \cite{Bian2026AirFMDDA}. In parallel, lightweight encoders, windowed attention, and patch-based processing provide additional mechanisms for limiting sequence length and attention cost. Collectively, these architectural directions indicate a shift toward computation-aware WFMs that balance representation quality, latency, and memory requirements for practical deployment in AI-native 6G systems.
Table~\ref{tab:architectural_tradeoffs} summarizes the principal architectural trade-offs among representative wireless foundation model designs, highlighting the balance between representation capability, computational efficiency, and deployment suitability across different wireless scenarios.
\begin{table*}[t]
\centering
\caption{Scalability Comparison of Representative Wireless Foundation Model Architectures}
\label{tab:architectural_tradeoffs}
\renewcommand{\arraystretch}{1.95}
\setlength{\tabcolsep}{4pt}
\scriptsize

\begin{tabular}{|p{2.5cm}|p{2.5cm}|p{3.7cm}|p{3.8cm}|p{3.4cm}|}
\hline

\textbf{Architecture Category}
&
\textbf{Representative Models}
&
\textbf{Scalability Advantages}
&
\textbf{Main Limitations}
&
\textbf{Typical Deployment}
\\
\hline

Transformer / Vision Transformer
&
WirelessGPT, LWM, WavesFM
&
Excellent representation learning; strong multi-task transfer; captures long-range spatial and temporal dependencies
&
Large parameter count; quadratic self-attention complexity; high memory consumption
&
Cloud servers, base stations, GPU-enabled edge nodes
\\
\hline

Masked Autoencoder
&
WavesFM, WiFo, ContraWiMAE
&
Label-efficient pre-training; learns hidden channel structures; improves data efficiency
&
Sensitive to masking strategy; requires diverse pre-training data
&
Channel estimation, CSI feedback, channel prediction
\\
\hline

Prompt-guided Encoder--Decoder
&
MuSE-FM
&
Supports heterogeneous task formats; enables efficient task adaptation
&
Additional architectural complexity; prompt design remains challenging
&
Multi-task wireless intelligence
\\
\hline

Parameter-Efficient Adaptation
&
LoRA, Adapter-based WFMs
&
Very small number of trainable parameters; low adaptation cost; efficient model updates
&
Performance may degrade under highly mismatched domains
&
Edge adaptation, continual learning, model updates
\\
\hline

Lightweight Encoders
&
Tiny-WiFo, Lightweight Time-Series FM
&
Low latency; reduced memory footprint; suitable for embedded devices
&
Limited global context modeling
&
IoT devices, mobile terminals, real-time inference
\\
\hline

State-Space / Mamba
&
WiMamba, ComHymba
&
Near-linear computational complexity; efficient long-sequence modeling
&
Immature ecosystem; limited wireless benchmarks
&
Long CSI sequences, real-time wireless inference
\\
\hline

Domain-informed Attention
&
AirFM-DDA, ComHymba
&
Exploits wireless-domain priors; lower attention complexity
&
Requires domain-specific preprocessing
&
Large CSI tensors, physical-layer tasks
\\
\hline

Compression Techniques
&
Tiny Federated WFM, Quantized WFMs
&
Reduces model size, memory, and energy consumption
&
Potential accuracy degradation after compression
&
Resource-constrained edge devices and federated learning
\\
\hline

Multimodal Foundation Models
&
Multimodal WFM, WavesFM, MuSE-FM
&
Jointly learns communication, sensing, and localization representations
&
Higher computational and memory requirements due to modality fusion
&
ISAC, AI-native 6G, semantic communications
\\
\hline

\end{tabular}
\end{table*}

While each architectural family offers unique advantages, no single architecture is universally optimal for all wireless applications. Transformer-based models provide the strongest representation learning capability and multi-task transferability by capturing long-range spatial and temporal dependencies, but their quadratic self-attention complexity limits efficient deployment on resource-constrained devices. CNN-based architectures remain attractive for applications requiring low computational complexity and real-time inference, although they are less effective in modeling global wireless dependencies. Physics-informed architectures improve data efficiency, interpretability, and robustness by incorporating communication-domain knowledge into the learning process, but they may sacrifice flexibility when operating under highly diverse propagation environments. Multimodal architectures further enhance representation quality by jointly processing heterogeneous wireless observations; however, this comes at the cost of increased computational complexity and larger data requirements. Consequently, the choice of architecture should be guided by the target application, available computational resources, and deployment constraints rather than by representation accuracy alone.

\begin{table*}[t]
\centering
\caption{Comparison of major wireless foundation model architectures.}
\label{tab:architecture_comparison}

\renewcommand{\arraystretch}{1.7}

\begin{tabular}{lccccc}
\hline
\textbf{Architecture} &
\textbf{Representation} &
\textbf{Complexity} &
\textbf{Data Requirement} &
\textbf{Edge Deployment} &
\textbf{Generalization} \\
\hline

Transformer-based &
Excellent &
High &
High &
Moderate &
Excellent \\

CNN-based &
Good &
Low &
Moderate &
Excellent &
Moderate \\

Physics-informed &
Good &
Moderate &
Low &
Good &
High \\

Multimodal &
Excellent &
Very High &
Very High &
Limited &
Excellent \\

\hline
\end{tabular}
\end{table*}
Table~\ref{tab:architecture_comparison} highlights that the architectural design of a wireless foundation model involves balancing representation capability, computational efficiency, data availability, and deployment requirements. Transformer-based models are well suited for large-scale pre-training and multi-task learning, whereas CNN-based and physics-informed architectures remain attractive for latency-sensitive and resource-constrained applications. Multimodal architectures provide the richest representations for AI-native 6G systems but require significantly larger datasets and computational resources. Future wireless foundation models are therefore expected to combine the strengths of multiple architectural paradigms rather than relying on a single backbone.

\section{Pre-Training Strategies}
\subsection{Self-Supervised Learning}

SSL has become the dominant pre-training strategy for wireless foundation models because it enables representation learning from large volumes of unlabeled wireless data. Unlike supervised learning, which depends on manually annotated datasets, SSL generates supervisory signals directly from the input data through carefully designed pretext tasks. This paradigm is particularly attractive for wireless communications, where raw measurements such as CSI, IQ samples, CIR, and spectrograms are readily available, whereas obtaining accurate labels is expensive and time-consuming \cite{Jing2020SelfSupervised,Liu2021SelfSupervisedSurvey}.
The primary objective of SSL is to learn generalized wireless representations that capture the spatial, temporal, and frequency-domain characteristics of radio signals. After large-scale pre-training, these representations can be efficiently transferred to downstream tasks, including channel estimation, channel prediction, signal detection, beam prediction, localization, and wireless sensing, using only limited labeled data. Consequently, SSL significantly reduces annotation costs while improving robustness and generalization across diverse wireless environments.

Among the various SSL objectives, masked signal modeling has emerged as one of the most effective approaches for wireless foundation models. Inspired by masked language modeling and masked image modeling, portions of the wireless input are intentionally hidden, and the model is trained to reconstruct the missing information. Depending on the application, the masked regions may correspond to CSI elements, IQ samples, OFDM resource grids, or time-frequency patches. By reconstructing these missing measurements, the model learns the intrinsic spatial, temporal, and spectral structure of wireless signals without requiring explicit supervision. Representative examples include WavesFM and scalable masked channel models, which employ reconstruction-based pre-training for channel estimation, CSI feedback, and channel prediction \cite{He2022MAE,Guo2026MaskedChannelModel}.
Another widely adopted SSL objective is contrastive representation learning, which learns discriminative feature embeddings by maximizing the similarity between different augmented views of the same wireless sample while separating unrelated samples in the latent space. Positive pairs are typically generated through signal augmentations or multiple observations of the same propagation environment, whereas negative pairs correspond to unrelated wireless measurements.

This objective encourages invariant representation learning and improves robustness to noise, mobility, and distribution shifts. Recent studies have demonstrated its effectiveness for CSI representation learning, localization, and RF signal classification \cite{Chen2020SimCLR,Grill2020BYOL}.
Rather than relying on a single objective, contemporary wireless foundation models frequently combine masked reconstruction and contrastive learning during pre-training. Reconstruction objectives encourage the model to capture the structural characteristics of wireless signals, whereas contrastive objectives improve representation discrimination and transferability. This combination has become the prevailing pre-training strategy for modern wireless foundation models.
\begin{figure*}[t]
    \centering
    \includegraphics[width=\textwidth]{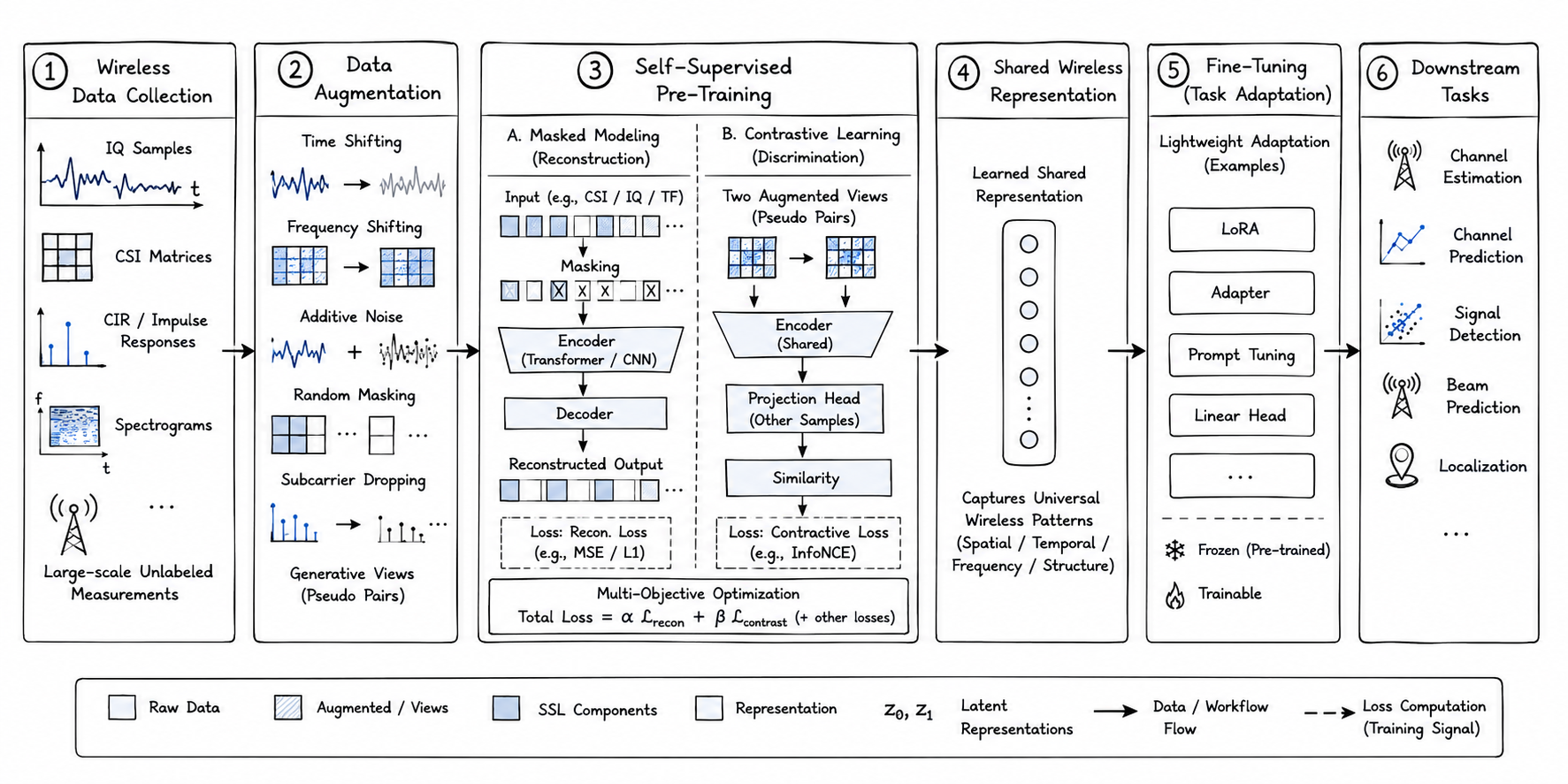}
    \caption{Self-supervised pre-training workflow for wireless foundation models.}
    \label{fig:ssl_pretraining}
\end{figure*}

As illustrated in Fig.~\ref{fig:ssl_pretraining}, heterogeneous wireless measurements are first collected and transformed through signal-domain augmentation to generate multiple views of the same wireless observation. The augmented data are then used to optimize complementary self-supervised objectives, including masked signal reconstruction and contrastive representation learning, enabling the model to capture both structural and discriminative characteristics of wireless signals. The resulting shared wireless representation forms the pre-trained backbone of the foundation model and can subsequently be adapted to downstream tasks through lightweight techniques such as LoRA, adapters, prompt tuning, or task-specific prediction heads. This unified pre-training paradigm substantially reduces the dependence on labeled datasets while improving scalability and transferability across a broad range of wireless communication applications.

\subsection{Data Generation and Simulation-Based Pre-Training}

The effectiveness of wireless foundation models depends critically on the availability of large-scale and diverse pre-training datasets. However, collecting real-world wireless measurements is expensive, time-consuming, and often constrained by deployment cost, hardware availability, and privacy considerations. Furthermore, constructing labeled datasets for applications such as channel estimation, beam prediction, and localization requires extensive measurement campaigns under different propagation environments, antenna configurations, and mobility conditions. Consequently, synthetic data generation has become an indispensable component of the pre-training pipeline for wireless foundation models \cite{molisch2012wireless,Rappaport2015Millimeter}.

Simulation-based pre-training relies on realistic wireless channel simulators to generate large numbers of channel realizations under diverse propagation conditions. Classical stochastic channel models, including Rayleigh, Rician, Nakagami-$m$, WINNER II, and the 3GPP spatial channel model, are widely used to emulate fading, path loss, shadowing, Doppler effects, and multipath propagation. These simulators produce CSI, CIR, and received signal samples over a broad range of SNRs, carrier frequencies, antenna configurations, and mobility scenarios, enabling foundation models to learn from considerably more diverse wireless environments than would be feasible through field measurements alone \cite{3GPP38901,Kyosti2007WinnerII}.
Deterministic ray-tracing has further enhanced the realism of synthetic wireless datasets by explicitly modeling electromagnetic wave propagation within site-specific environments. Unlike stochastic channel models, ray-tracing captures reflections, diffraction, scattering, blockage, and other propagation phenomena arising from the physical geometry of buildings and surrounding objects. Modern ray-tracing platforms therefore provide highly realistic datasets for millimeter-wave, terahertz, localization, beam management, and ISAC applications \cite{RemcomWirelessInSite,SionnaRT}.
Rather than relying exclusively on either simulated or measured data, many recent wireless foundation models adopt hybrid pre-training strategies that combine both sources. Synthetic datasets provide broad coverage across propagation conditions and communication scenarios, whereas real-world measurements reduce the simulation-to-reality (Sim2Real) gap and improve deployment robustness. By systematically varying channel models, antenna arrays, carrier frequencies, mobility patterns, hardware impairments, and interference conditions, simulation-based pre-training substantially improves representation diversity while reducing overfitting to a particular wireless environment\cite{sai2026machine}.

\begin{figure*}[t]
    \centering
    \includegraphics[width=\textwidth]{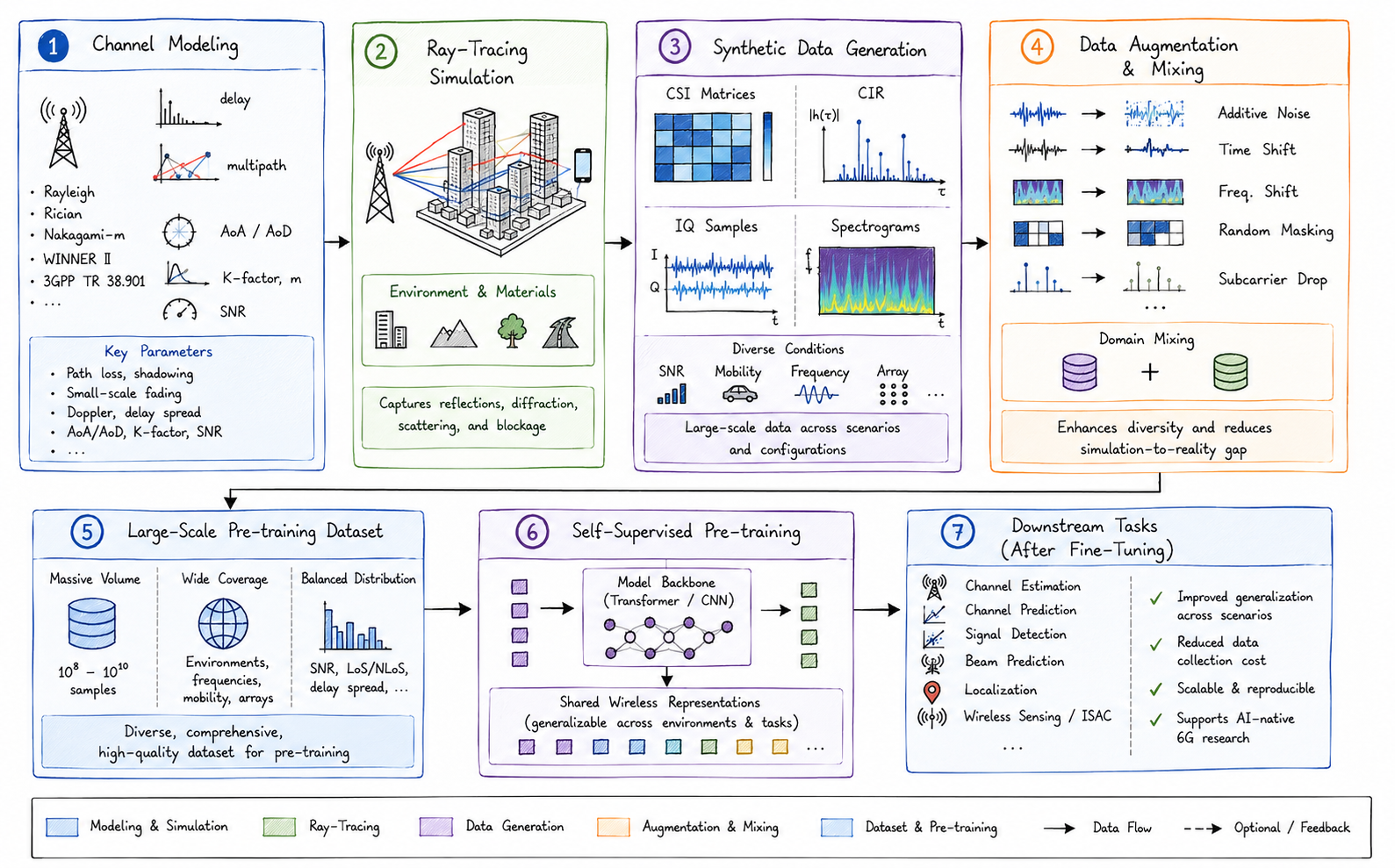}
    \caption{Simulation-based data generation and pre-training pipeline for wireless foundation models.}
    \label{fig:simulation_pretraining}
\end{figure*}

Despite these advantages, simulation-based pre-training also presents important challenges. Synthetic datasets inevitably simplify real propagation environments and cannot fully capture hardware non-idealities, environmental dynamics, or unexpected interference sources. Bridging the resulting Sim2Real gap remains an active research topic, motivating the development of higher-fidelity simulators, domain adaptation techniques, and hybrid datasets that combine simulation with real-world measurements\cite{alkhalifah2022mlreal}.
Figure~\ref{fig:simulation_pretraining} illustrates the general workflow of simulation-based pre-training for wireless foundation models.

As illustrated in Fig.~\ref{fig:simulation_pretraining}, the workflow begins with stochastic channel modeling and deterministic ray-tracing, which generate large-scale synthetic wireless measurements under configurable propagation environments. The generated CSI, IQ samples, CIR, and spectrograms are subsequently augmented and integrated with real-world measurements to form a comprehensive pre-training corpus. This hybrid dataset enables foundation models to learn generalized wireless representations that transfer effectively across different channel conditions, antenna configurations, frequency bands, and mobility scenarios. Following large-scale self-supervised pre-training, the learned representations can be efficiently adapted to downstream tasks, including channel estimation, channel prediction, beam prediction, localization, and integrated sensing, thereby providing a scalable and cost-effective alternative to relying solely on extensive real-world measurement campaigns.

\subsection{Transfer Learning and Fine-Tuning}

Transfer learning enables a pre-trained wireless foundation model to adapt its learned representations to new communication tasks using significantly less labeled data than training from scratch. Instead of independently developing models for channel estimation, signal detection, beam prediction, or localization, a single pre-trained backbone can be efficiently transferred to multiple downstream applications through task-specific adaptation. This paradigm substantially reduces training cost, improves data efficiency, and accelerates deployment across diverse wireless environments \cite{Pan2010TransferLearning}.

The most direct adaptation strategy is full fine-tuning, in which all model parameters are updated using task-specific data. Although this approach generally provides the highest task-specific performance, it also requires considerable computational resources, memory, and storage because a separate model must be maintained for each downstream application \cite{Howard2018UniversalLanguageModel}. Such requirements become increasingly impractical for AI-native 6G systems that are expected to support numerous communication and sensing tasks simultaneously.
To improve adaptation efficiency, recent research has focused on parameter-efficient fine-tuning (PEFT), where only a small subset of parameters is optimized while the majority of the pre-trained backbone remains fixed. Representative techniques include LoRA, adapters, prefix tuning, and bias-only optimization. These methods significantly reduce computational complexity, communication overhead, and memory consumption while preserving most of the performance achieved by full fine-tuning \cite{Houlsby2019Adapters,Lester2021PromptTuning}. For example, WavesFM adopts LoRA modules to efficiently adapt a shared backbone across communication, sensing, and localization tasks without maintaining multiple independently trained models \cite{Aboulfotouh2025WavesFM}.

\subsection{Prompting and In-Context Learning}

Prompt-based learning has recently emerged as an alternative adaptation strategy that further reduces the computational cost of transferring foundation models to downstream tasks. Instead of modifying the model parameters, prompt tuning introduces a small set of learnable prompt embeddings or task instructions that guide the pre-trained model toward a target application while preserving the knowledge acquired during pre-training \cite{Liu2023PromptSurvey}. Representative approaches include soft prompt tuning, which learns continuous prompt embeddings, and prefix tuning, which injects trainable vectors into intermediate Transformer layers without updating the backbone parameters \cite{Li2021PrefixTuning,Lester2021PromptTuning}.
Another emerging paradigm is in-context learning, where a pre-trained model performs a new task by conditioning on a small number of demonstration examples rather than updating its parameters. Although originally proposed for large language models, this capability has attracted growing interest in wireless communications because it enables rapid adaptation to unseen channel conditions, network configurations, or communication objectives with little or no retraining \cite{Brown2020GPT3,Dong2024ICLSurvey}. Such flexibility is particularly attractive for AI-native wireless systems operating in highly dynamic environments.


These adaptation strategies represent a progression from computationally intensive retraining toward increasingly lightweight and flexible learning mechanisms. As wireless foundation models continue to evolve, parameter-efficient fine-tuning, prompt-based adaptation, and in-context learning are expected to become key technologies for enabling scalable multi-task wireless intelligence across heterogeneous AI-native 6G systems.

\section{Applications of Wireless Foundation Models}
\subsection{Channel Estimation and Equalization}

Channel estimation and equalization are among the most mature application domains of wireless foundation models because they directly determine the reliability, throughput, and spectral efficiency of wireless communication systems. Conventional deep learning approaches generally train separate models for specific channel models, antenna configurations, or SNR ranges. Consequently, their performance often degrades under unseen propagation environments, mobility conditions, or hardware configurations, requiring extensive retraining and limiting their scalability in practical wireless networks \cite{gadhafi2026machine}.

Wireless foundation models overcome these limitations by learning generalized channel representations through large-scale self-supervised pre-training. Instead of optimizing for a single propagation scenario, the pre-trained backbone captures common spatial, temporal, and frequency-domain characteristics shared across diverse wireless environments. These learned representations can subsequently be adapted to channel estimation, channel prediction, CSI feedback, and equalization using only limited task-specific supervision, thereby improving generalization while significantly reducing retraining cost \cite{jiang2025towards,Alikhani2025LWM}.

Recent studies demonstrate the effectiveness of this paradigm. WirelessGPT employs a Transformer-based backbone to learn reusable channel representations that support multiple physical-layer tasks, while the LWM exploits large-scale channel datasets to improve robustness across heterogeneous propagation environments. Similarly, WavesFM combines masked signal modeling with a Vision Transformer architecture to jointly support channel estimation, CSI feedback, localization, and sensing using a shared pre-trained model. More recently, channel foundation models have incorporated masked channel reconstruction and contrastive learning objectives to further improve robustness under channel variations and limited labeled data \cite{Guo2026MaskedChannelModel, perenda2023contrastive}.
Channel equalization has also benefited from transferable wireless representations. Rather than designing separate equalizers for individual channel conditions, wireless foundation models learn generalized propagation characteristics that enable more robust compensation for channel distortion before signal demodulation. Such shared representations improve equalization performance across varying propagation environments, mobility conditions, and hardware impairments while reducing the need for repeated model retraining\cite{shanmugam2026improving}. As wireless communication systems continue to evolve toward highly dynamic AI-native 6G networks, wireless foundation models are expected to provide unified physical-layer intelligence capable of jointly supporting channel estimation, equalization, prediction, CSI compression, and related signal processing tasks within a single scalable framework.
\subsection{Signal Detection and Modulation Recognition}

Signal detection and modulation recognition represent another important application domain of wireless foundation models. Accurate signal detection is essential for recovering transmitted symbols in MIMO and OFDM systems \cite{khurshid2026sormmse}, whereas automatic modulation recognition supports adaptive communication, spectrum monitoring, cognitive radio, and intelligent wireless network management. Conventional deep learning solutions typically train separate models for individual modulation formats, antenna configurations, or channel conditions, limiting their ability to generalize across heterogeneous wireless environments and communication standards \cite{Doha2025Receivers}.

Wireless foundation models address these limitations by learning generalized signal representations from large collections of unlabeled IQ samples, spectrograms, and channel measurements. Through large-scale self-supervised pre-training, the model captures invariant signal characteristics that remain robust under noise, fading, interference, and hardware impairments. These representations can subsequently be adapted to signal detection, modulation classification, wireless technology recognition, and RF fingerprinting using only limited task-specific supervision, substantially reducing the dependence on large labeled datasets \cite{Mashaal2025IQFM,Cheraghinia2025Unified}.
Representative studies demonstrate the effectiveness of this approach. IQFM learns universal feature representations directly from raw IQ streams, enabling efficient transfer across multiple wireless signal processing tasks. Similarly, unified wireless foundation models for wireless technology recognition and localization employ a shared pre-trained backbone that simultaneously supports signal classification and positioning without requiring independent task-specific networks. Compared with conventional deep learning models \cite{sadia2025irs}, these approaches exhibit improved robustness under unseen propagation environments because they learn reusable wireless representations rather than features tailored to a single dataset or communication scenario \cite{morocho2020breaking}.

Beyond traditional communication systems, wireless foundation models are expected to play an increasingly important role in cognitive radio, spectrum sensing, RF environment awareness, and intelligent network monitoring. Their ability to rapidly adapt to new modulation formats, communication protocols, and spectrum conditions makes them well suited for AI-native 6G systems, where communication environments evolve continuously and require scalable, transferable, and context-aware wireless intelligence.

\subsection{Beamforming and Resource Allocation}

Beamforming and resource allocation are fundamental components of modern wireless communication systems because they directly influence network capacity, spectral efficiency, energy efficiency, and QoS \cite{khan2025drl}. Conventional approaches formulate these problems as non-convex optimization tasks that require accurate CSI together with iterative optimization algorithms. As wireless networks evolve toward massive MIMO, millimeter-wave (mmWave) communications, ultra-dense deployments, and ISAC, solving these optimization problems in real time becomes increasingly challenging \cite{Bjornson2017MassiveMIMO,11429479,Shen2018LearningBeamforming}.
Wireless foundation models offer a scalable alternative by learning generalized representations of wireless channels, user mobility, traffic patterns, and network dynamics through large-scale pre-training. Instead of optimizing beamforming vectors or resource allocation policies independently for each deployment scenario, a shared pre-trained backbone can be efficiently adapted to different propagation environments, antenna configurations, and service requirements using lightweight adaptation techniques\cite{zhu2025wireless}. This paradigm reduces computational complexity while improving robustness and transferability across heterogeneous wireless networks.

Recent studies demonstrate the effectiveness of this approach. MuSE-FM jointly exploits wireless measurements and environmental context to support multi-task communication optimization, while wireless foundation models for multi-task prediction learn shared representations for traffic forecasting, mobility prediction, and resource allocation. More recently, large AI models have explored intent-driven wireless optimization by incorporating natural-language reasoning into resource management and autonomous network control, providing a new direction for intelligent radio access networks \cite{luo2026ai,Sheng2025WFMTaskPrediction,Huang2026LargeAI}.
Compared with conventional optimization methods that solve individual problems independently, wireless foundation models enable unified decision-making across multiple network functions. By jointly optimizing beamforming, power allocation, interference mitigation, spectrum management, and user scheduling within a single pre-trained framework, they provide a scalable foundation for adaptive and autonomous wireless resource management. This capability is expected to become increasingly important as AI-native 6G systems evolve toward fully self-optimizing communication networks.

\subsection{Localization and Integrated Sensing}

Wireless localization and integrated sensing have emerged as key application domains for wireless foundation models because future 6G networks are expected to simultaneously provide high-precision positioning, environmental perception, and reliable communication through ISAC \cite{localization}. Achieving these capabilities requires the joint processing of heterogeneous wireless measurements, including CSI, CIR, radar echoes, and IQ signals, under dynamic propagation environments characterized by mobility, blockage, and NLOS conditions \cite{Wymeersch2025ISAC,Liu2024ISACSurvey}.
Wireless foundation models address these challenges by learning unified representations from multimodal wireless observations through self-supervised pre-training. Rather than developing separate models for localization, sensing, and communication, a shared backbone captures common spatial, temporal, and propagation characteristics that can be efficiently adapted to multiple downstream tasks. This unified learning paradigm improves robustness across diverse environments while reducing the dependence on large annotated positioning datasets.

Recent studies demonstrate the effectiveness of this approach. WavesFM employs a Vision Transformer backbone with parameter-efficient adaptation to jointly support communication, localization, and sensing. Similarly, multimodal wireless foundation models combine CSI, IQ signals, radar measurements, and environmental information to enhance positioning accuracy and sensing performance. More recent frameworks further integrate multimodal sensing and communication within a unified architecture, enabling joint localization, environmental perception, and channel prediction using a single pre-trained model \cite{Aboulfotouh2025WavesFM,yu2025channelgpt}.
The integration of localization and sensing within a common foundation model represents an important step toward context-aware wireless intelligence. By jointly exploiting heterogeneous sensing modalities, wireless foundation models can support applications such as autonomous driving, digital twins, smart manufacturing, and intelligent transportation systems, where communication, positioning, and environmental awareness must operate seamlessly. This unified capability is expected to become a core building block of AI-native 6G networks, enabling scalable and adaptive ISAC services across diverse deployment scenarios.

\subsection{Semantic and Goal-Oriented Communications}

Semantic and goal-oriented communications have emerged as key paradigms for next-generation wireless networks by shifting the communication objective from faithfully transmitting every bit to delivering information that is most relevant for a target task \cite{guo2023semantic}. Unlike conventional communication systems that optimize metrics such as bit error rate (BER), throughput, or spectral efficiency, semantic communications focus on preserving the meaning of transmitted information, whereas goal-oriented communications further optimize transmission according to the receiver's task or decision objective \cite{Qin2022Semantic, Khan2026Semantic}.
Wireless foundation models provide a natural framework for this paradigm because they learn high-level representations that capture relationships among wireless signals, environmental context, user behavior, and network states. Rather than processing communication signals independently, a shared pre-trained model can jointly encode semantic information across heterogeneous modalities, enabling more efficient compression, robust semantic encoding, and adaptive decision-making while reducing unnecessary wireless transmissions.

Recent studies have demonstrated the potential of large foundation models for semantic-aware wireless systems. large language models (LLMs) have been explored for user-intent understanding, semantic reasoning, and autonomous network management, while multimodal wireless foundation models combine wireless measurements with vision, radar, and environmental information to support semantic communication and integrated sensing \cite{khurshid2026llm}. These capabilities improve communication efficiency in applications such as autonomous driving, intelligent transportation, industrial automation, extended reality (XR), and digital twins, where successful task completion is often more important than accurate symbol recovery \cite{Liang2026LLMWireless,Jiang2026LargeAI, 11104471}.
Although semantic wireless foundation models remain at an early stage of development, they represent an important evolution beyond conventional physical-layer optimization. Future research will require unified semantic representations for wireless data, evaluation metrics that quantify task effectiveness rather than bit-level accuracy, and closer integration between semantic reasoning and physical-layer signal processing\cite{islam2024deep}. These advances are expected to establish semantic and goal-oriented communications as one of the defining applications of wireless foundation models in AI-native 6G systems.

The representative applications discussed above demonstrate that wireless foundation models are evolving from specialized physical-layer solutions into unified learning frameworks capable of supporting communication, sensing, localization, and network optimization. Despite sharing the common objective of learning reusable wireless representations, existing models differ considerably in their architectures, pre-training strategies, adaptation mechanisms, supported protocol layers, and target applications. Table~\ref{tab:wfm_comparison} summarizes representative wireless foundation models and highlights the key design trends driving this evolution.

\begin{table*}[t]
\centering
\caption{Comparison of Representative Wireless Foundation Models}
\label{tab:wfm_comparison}
\renewcommand{\arraystretch}{1.25}
\resizebox{\textwidth}{!}{
\begin{tabular}{|p{2.5cm}|p{1.8cm}|p{2.2cm}|p{2.2cm}|p{1.4cm}|p{2.6cm}|p{2.2cm}|p{1.8cm}|p{3.8cm}|}
\hline
\textbf{Foundation Model}
&
\textbf{Architecture}
&
\textbf{Pre-training Strategy}
&
\textbf{Dataset}
&
\textbf{Wireless Layer}
&
\textbf{Application}
&
\textbf{Transfer Learning}
&
\textbf{Evaluation Metric}
&
\textbf{Main Contribution}
\\
\hline

WirelessGPT \cite{Yang2025WirelessGPT}
&
Transformer
&
Self-supervised
&
CSI, IQ
&
PHY
&
Channel estimation, detection
&
Fine-tuning
&
NMSE, BER
&
General-purpose Transformer backbone for multi-task wireless communications.
\\
\hline

Large Wireless Model (LWM) \cite{Alikhani2025LWM}
&
Transformer
&
SSL
&
Large wireless channel dataset
&
PHY
&
Channel prediction
&
Fine-tuning
&
NMSE
&
Learns transferable channel representations across propagation environments.
\\
\hline

WavesFM \cite{Aboulfotouh2025WavesFM}
&
Vision Transformer
&
Masked Modeling
&
CSI, IQ, Spectrograms
&
PHY
&
Channel estimation, localization, sensing
&
LoRA
&
NMSE, Localization Error
&
Unified foundation model supporting communication, sensing, and localization.
\\
\hline

MuSE-FM \cite{Sheng2025WFMTaskPrediction}
&
Transformer
&
Multi-modal SSL
&
CSI + Environmental Data
&
Cross-layer
&
Beamforming, Resource Allocation
&
LoRA
&
Spectral Efficiency
&
Environment-aware multi-task optimization framework.
\\
\hline

WiFo \cite{Liu2025WiFo}
&
Transformer
&
Masked Channel Modeling
&
CSI
&
PHY
&
Channel prediction
&
Fine-tuning
&
NMSE
&
Long-term channel prediction using pre-trained representations.
\\
\hline

CSI2Vec \cite{Palhares2025CSI2Vec}
&
Transformer Encoder
&
Contrastive Learning
&
CSI
&
PHY
&
Localization
&
Fine-tuning
&
Position Error
&
Universal CSI embedding for localization and channel charting.
\\
\hline

IQFM \cite{Mashaal2025IQFM}
&
Transformer
&
Self-supervised
&
IQ Streams
&
PHY
&
Signal Detection
&
Fine-tuning
&
BER
&
Learns transferable representations directly from raw IQ samples.
\\
\hline

Unified WFM \cite{Cheraghinia2025Unified}
&
Transformer
&
SSL
&
Wireless Technology Dataset
&
PHY
&
Technology Recognition
&
Fine-tuning
&
Classification Accuracy
&
Joint wireless technology recognition and localization.
\\
\hline

Multimodal WFM \cite{Aboulfotouh2025Multimodal}
&
Multi-modal Transformer
&
Multi-modal SSL
&
CSI, IQ, Radar
&
Cross-layer
&
ISAC, Localization
&
LoRA
&
Detection Accuracy
&
Joint communication and sensing representation learning.
\\
\hline

WiFo-MISAC \cite{Liu2026WiFoMiSAC}
&
Transformer
&
Multi-modal SSL
&
Wireless Sensing Dataset
&
Cross-layer
&
Integrated Sensing and Communication
&
LoRA
&
Detection Accuracy
&
Unified multimodal sensing and communication foundation model.
\\
\hline

AirFM-DDA \cite{Bian2026AirFM}
&
Physics-informed Transformer
&
Self-supervised
&
Delay--Doppler Channel Dataset
&
PHY
&
Channel Estimation
&
Prompt / Fine-tuning
&
NMSE
&
Introduces physics-aware positional encoding for wireless foundation models.
\\
\hline

\end{tabular}}
\end{table*}

Table~\ref{tab:wfm_comparison} reveals several important trends in the current development of wireless foundation models.
First, Transformer-based architectures have become the dominant backbone owing to their ability to model long-range spatial and temporal dependencies while supporting scalable multi-task learning. Although convolutional and physics-informed architectures remain valuable for specific applications, most recent foundation models favor Transformer variants because of their flexibility across heterogeneous wireless modalities.

Second, self-supervised learning has emerged as the predominant pre-training paradigm. Techniques such as masked modeling, contrastive learning, and multimodal representation learning effectively exploit the abundance of unlabeled wireless measurements, reducing the dependence on costly annotated datasets while improving model generalization across diverse deployment scenarios.

Third, adaptation strategies are shifting from conventional full-parameter fine-tuning toward parameter-efficient methods, particularly  LoRA. These lightweight techniques significantly reduce computational and memory overhead while enabling a single pre-trained backbone to support multiple downstream communication and sensing tasks.

Finally, the scope of wireless foundation models is expanding beyond traditional physical-layer applications. While early studies primarily focused on channel estimation, prediction, and signal detection, more recent models increasingly support beamforming, resource allocation, localization, ISAC, semantic communications, and cross-layer network intelligence. This evolution reflects a broader transition from task-specific deep learning models toward unified wireless intelligence capable of serving diverse communication, sensing, and network management functions within a common foundation-model framework.


\section{Datasets, Benchmarks, and Evaluation Metrics}

\subsection{Public Wireless Datasets}

Large-scale datasets are the cornerstone of wireless foundation models because they provide the diverse observations required for large-scale pre-training, benchmarking, and downstream adaptation. Unlike conventional supervised learning, which relies on task-specific labeled datasets, wireless foundation models benefit from heterogeneous collections of CSI, IQ samples, CIR, spectrograms, localization measurements, sensing observations, and network telemetry collected under diverse propagation environments\cite{ying2026specialist}. The diversity of these datasets enables the learning of transferable wireless representations that generalize across communication tasks, deployment scenarios, and network configurations.

Several public datasets have become standard benchmarks for wireless AI research. DeepMIMO is one of the most widely adopted ray-tracing-based datasets for millimeter-wave and massive MIMO communications. Generated using Remcom Wireless InSite, it provides configurable channel measurements suitable for beam prediction, channel estimation, localization, and deep learning-based wireless communications \cite{Alkhateeb2019DeepMIMO}. Raymobtime further extends this concept by incorporating user mobility, making it particularly valuable for evaluating beam tracking and channel prediction algorithms under dynamic propagation conditions \cite{Raymobtime2018}.

Public datasets have also been developed for signal recognition and multimodal wireless intelligence. RadioML provides labeled IQ samples covering multiple modulation formats across different SNRs, making it one of the most widely used benchmarks for automatic modulation recognition and RF signal classification \cite{OShea2018RadioML}. More recently, DeepSense~6G has introduced a multimodal benchmark that combines wireless communication, sensing, localization, and environmental information, reflecting the growing emphasis on AI-native 6G applications\cite{DeepSense6G2024}. Similarly, ViWi integrates wireless channel measurements with visual information to support vision-aided beamforming and localization \cite{ViWi2020}, while Sionna RT generates physics-based synthetic datasets through differentiable ray tracing for wireless channel modeling and sensing applications \cite{SionnaRT2023}.

Despite their importance, current wireless datasets remain significantly smaller and less diverse than the datasets used to train foundation models in natural language processing and computer vision. Most existing benchmarks focus on specific communication tasks, frequency bands, or propagation environments, limiting their ability to support truly general-purpose wireless intelligence. 
Future wireless foundation models will therefore require large-scale multimodal datasets that combine synthetic channel simulations, real-world measurements, sensing observations, environmental context, and network management information under standardized benchmark protocols. Such datasets will play a critical role in improving model generalization, enabling fair comparison among wireless foundation models, and accelerating the development of scalable AI-native 6G systems. Table~\ref{tab:datasets} summarizes representative public datasets currently used for the development and evaluation of wireless foundation models.

\begin{table*}[t]
\centering
\caption{Representative Public Datasets for Wireless Foundation Models}
\label{tab:datasets}
\renewcommand{\arraystretch}{1.2}
\setlength{\tabcolsep}{4pt}
\scriptsize
\begin{tabular}{|p{2.0cm}|p{2.0cm}|p{2.3cm}|p{3.0cm}|p{2.0cm}|p{4.0cm}|}
\hline
\textbf{Dataset} &
\textbf{Modality} &
\textbf{Source} &
\textbf{Representative Applications} &
\textbf{Typical Metrics} &
\textbf{Key Features} \\
\hline

DeepMIMO \cite{Alkhateeb2019DeepMIMO}
&
CSI
&
Ray tracing
&
Beam prediction, channel estimation, localization
&
NMSE, Spectral Efficiency
&
Configurable mmWave channel dataset with multiple scenarios.
\\
\hline

Raymobtime \cite{Raymobtime2018}
&
CSI
&
Ray tracing + Mobility
&
Beam tracking, channel prediction
&
NMSE
&
Captures temporal channel evolution under user mobility.
\\
\hline

RadioML \cite{OShea2018RadioML}
&
IQ Samples
&
Simulation
&
Modulation recognition, RF classification
&
Accuracy
&
Widely adopted benchmark for wireless signal classification.
\\
\hline

DeepSense~6G \cite{DeepSense6G2024}
&
CSI, Radar, Sensor Data
&
Real + Simulation
&
Localization, ISAC
&
Localization Error, Detection Accuracy
&
Large-scale multimodal benchmark for AI-native 6G.
\\
\hline

ViWi \cite{ViWi2020}
&
CSI + Images
&
Ray tracing
&
Vision-aided beamforming, localization
&
Beam Prediction Accuracy
&
Combines wireless channels with visual information.
\\
\hline

Sionna RT \cite{SionnaRT2023}
&
CSI, CIR
&
Differentiable Ray Tracing
&
Channel estimation, localization, sensing
&
NMSE
&
Physics-aware synthetic dataset generation framework.
\\
\hline

\end{tabular}
\end{table*}

\subsection{Performance Metrics}

Evaluating wireless foundation models requires a broader set of metrics than those traditionally used in wireless communications. Unlike conventional communication algorithms that are optimized for a single task, wireless foundation models are expected to support multiple downstream applications while maintaining strong generalization, computational efficiency, and adaptability across heterogeneous wireless environments\cite{yang2026generative}. Consequently, their evaluation should simultaneously consider communication performance, representation quality, transferability, and deployment efficiency.
For physical-layer applications, the most commonly adopted metrics remain the BER and the normalized mean square error (NMSE). BER quantifies the reliability of signal detection by measuring the proportion of incorrectly detected bits, whereas NMSE evaluates the accuracy of estimated channel coefficients relative to the ground truth. Lower BER and NMSE values generally indicate improved communication reliability and channel estimation performance \cite{Goldsmith2005Wireless}.

Higher-layer applications are typically evaluated using network-oriented metrics, including spectral efficiency, energy efficiency, throughput, latency, and localization error. Spectral efficiency, usually expressed in bits/s/Hz, measures the effectiveness of spectrum utilization, whereas localization accuracy is commonly quantified using the RMSE or the mean positioning error. For ISAC, additional sensing metrics such as detection probability, false alarm rate, and sensing accuracy are widely adopted \cite{Liu2024ISACSurvey}.
Beyond application-specific performance, wireless foundation models must also be evaluated according to their ability to generalize across diverse deployment scenarios. 

Unlike conventional deep learning models, foundation models are expected to transfer knowledge across different propagation environments, carrier frequencies, antenna configurations, mobility conditions, and communication tasks. Consequently, cross-domain evaluation, few-shot adaptation, transfer learning performance, and out-of-distribution (OOD) robustness have become essential benchmark criteria for measuring the effectiveness of transferable wireless representations\cite{xie2026learning}.
Computational efficiency represents another critical evaluation dimension because wireless foundation models are expected to operate across cloud servers, edge platforms, and resource-constrained user devices. Common efficiency metrics include the number of trainable parameters, floating-point operations (FLOPs), inference latency, memory consumption, and energy usage\cite{huang2026edge}. In addition, parameter-efficient adaptation methods, such as LoRA and adapter-based tuning, are often assessed by comparing the percentage of trainable parameters and computational overhead relative to conventional full fine-tuning\cite{zhu2026text}.

Table~\ref{tab:metrics} summarizes the most commonly adopted evaluation metrics for wireless foundation models, together with their primary objectives and representative application domains.

\begin{table}[t]
\centering
\caption{Common Evaluation Metrics for Wireless Foundation Models}
\label{tab:metrics}
\renewcommand{\arraystretch}{1.2}
\scriptsize
\begin{tabular}{|p{2.3cm}|p{3.4cm}|p{2.2cm}|}
\hline
\textbf{Metric} &
\textbf{Purpose} &
\textbf{Representative Applications} \\
\hline

BER &
Measures signal detection reliability. Lower values indicate better communication performance. &
Signal detection, decoding \\
\hline

NMSE &
Evaluates channel estimation accuracy relative to the ground truth. &
Channel estimation, CSI prediction \\
\hline

Spectral Efficiency &
Measures throughput per unit bandwidth (bits/s/Hz). &
Beamforming, resource allocation \\
\hline

Localization Error &
Measures positioning accuracy using RMSE or mean error. &
Localization, ISAC \\
\hline

OOD Accuracy &
Evaluates robustness under unseen deployment environments. &
Generalization assessment \\
\hline

Transfer Learning Performance &
Measures adaptation capability using limited labeled data. &
Few-shot adaptation \\
\hline

Inference Latency &
Evaluates real-time deployment capability. &
Edge intelligence, online inference \\
\hline

FLOPs / Parameters &
Measures computational complexity and model efficiency. &
Architecture comparison \\
\hline

Memory Consumption &
Measures storage and deployment requirements. &
Edge devices, IoT \\
\hline

Energy Consumption &
Measures computational energy efficiency. &
Green AI, edge deployment \\
\hline

\end{tabular}
\end{table}

\subsection{Generalization and Robustness Evaluation}

Unlike conventional deep learning models that are typically evaluated under fixed training and testing conditions, wireless foundation models are expected to operate across diverse propagation environments, network configurations, hardware platforms, and communication standards. Consequently, evaluating only application-specific metrics such as BER or NMSE is insufficient\cite{pang2026beyond}. A comprehensive evaluation must also assess the model's ability to generalize across unseen wireless scenarios while maintaining robustness against distribution shifts, environmental variations, and adversarial perturbations.

Generalization evaluation measures how effectively a pre-trained wireless foundation model transfers to deployment conditions that differ from those encountered during pre-training. Representative evaluation scenarios include changes in SNR, carrier frequency, antenna configuration, user mobility, propagation environment, and hardware impairments. Since wireless channels are inherently dynamic, cross-domain evaluation has become an essential benchmark for measuring transferability across heterogeneous communication environments. Consequently, transfer learning performance, few-shot adaptation, and cross-scenario generalization are increasingly adopted as key evaluation criteria for wireless foundation models \cite{Akrout2023DomainGeneralization,Nguyen2023TransferLearning}.
OOD testing provides a complementary assessment by intentionally separating the training and testing distributions. Typical OOD scenarios include previously unseen propagation environments, different antenna arrays, new modulation formats, changing mobility patterns, and hardware variations. Models that maintain stable performance under these conditions are considered more suitable for practical AI-native wireless systems, where operating conditions continuously evolve \cite{Liu2021OOD,Hendrycks2021OOD}.

Robustness evaluation further examines the resilience of wireless foundation models against noise, interference, adversarial attacks, and hardware non-idealities. Adversarial robustness is commonly evaluated using attack methods such as the Fast Gradient Sign Method (FGSM) and  PGD, together with measurements of performance degradation under different perturbation strengths. In practical wireless deployments, robustness is also assessed under channel estimation errors, synchronization offsets, phase noise, nonlinear hardware distortion, quantization effects, and imperfect channel state information, providing a more realistic measure of deployment reliability \cite{Goodfellow2015FGSM,Madry2018PGD}.
Although considerable progress has been achieved, standardized evaluation protocols for wireless foundation models remain limited. Existing studies often employ different datasets, channel models, and experimental settings, making direct comparison difficult. Future benchmark suites should therefore integrate multi-domain datasets, standardized OOD evaluation, adversarial robustness testing, continual learning scenarios, and computational efficiency metrics to provide a comprehensive and reproducible assessment of wireless foundation models.
\section{Open Challenges and Research Directions}
\subsection{Data Scarcity and Domain Shift}

The development of wireless foundation models is fundamentally constrained by the availability of large-scale, diverse, and representative wireless datasets. Unlike natural language processing and computer vision, where foundation models are trained using billions of publicly available text documents and images, wireless communications lack standardized datasets of comparable scale and diversity\cite{hussain2026rise}. Existing public datasets are typically collected for specific communication tasks, propagation environments, carrier frequencies, or hardware platforms, limiting their suitability for learning truly general-purpose wireless representations. Consequently, many current wireless foundation models rely heavily on synthetic datasets generated using stochastic channel models or ray-tracing simulators, which cannot fully capture the complexity of real-world wireless environments \cite{wang2026tutorial}.

An even more fundamental challenge arises from the non-stationary nature of wireless environments. Wireless channels continuously evolve due to user mobility, environmental dynamics, network reconfiguration, hardware impairments, and spectrum utilization. As a result, the statistical distribution of wireless data changes over time, violating the independent and identically distributed (IID) assumption underlying most current pre-training strategies\cite{yue2026review}. Although large-scale pre-training improves representation learning, it cannot completely eliminate the performance degradation caused by distribution shifts between the pre-training and deployment environments. For example, a model trained on urban sub-6~GHz channels may perform poorly when deployed in millimeter-wave, terahertz, satellite, or industrial communication scenarios without additional adaptation.

This challenge distinguishes wireless foundation models from their counterparts in natural language processing and computer vision. While linguistic structures and visual features remain relatively stable across domains, wireless signals are governed by physical propagation mechanisms, antenna configurations, operating frequencies, and hardware characteristics that vary substantially across deployment scenarios. Consequently, simply increasing dataset size is insufficient for achieving robust generalization. Future wireless foundation models must instead learn representations that remain invariant to environmental changes while preserving information relevant to downstream communication tasks.

Current research primarily addresses domain shift through transfer learning, domain adaptation, and parameter-efficient fine-tuning. Although these approaches improve adaptation efficiency, they remain reactive because they require additional data from the target domain after deployment. A more scalable research direction is proactive domain generalization, where the pre-training objective explicitly encourages invariant representation learning across diverse propagation environments before deployment. Integrating self-supervised learning with physics-informed constraints, causal representation learning, and meta-learning may further improve the ability of wireless foundation models to generalize across previously unseen scenarios \cite{11549919, Nguyen2023TransferLearning}.
Another critical challenge is the lack of standardized benchmark datasets and evaluation protocols. Existing studies employ different channel models, simulation platforms, antenna configurations, mobility patterns, and experimental settings, making direct comparison among wireless foundation models difficult. Similar to the role of ImageNet in computer vision, the wireless community requires open, large-scale benchmark datasets that integrate synthetic simulations, real-world measurements, multimodal sensing observations, and network management information under standardized evaluation methodologies. Such benchmarks would not only enable fair comparison among competing models but also accelerate the development of reproducible and scalable wireless foundation models.

Looking ahead, future dataset development should extend beyond communication signals alone. AI-native wireless networks increasingly generate heterogeneous information, including CSI, IQ samples, radar observations, LiDAR data, environmental maps, mobility traces, traffic statistics, and network telemetry. Constructing multimodal datasets that jointly capture communication, sensing, localization, and networking information will therefore be a key prerequisite for developing truly general-purpose wireless foundation models capable of supporting diverse wireless intelligence tasks within a unified learning framework.

\subsection{Interpretability and Physical Consistency}

The remarkable performance of wireless foundation models has been driven by increasingly expressive neural architectures, particularly large Transformer-based models. However, this improved representation capability is accompanied by reduced interpretability. Unlike conventional communication algorithms, whose behavior can be explained through estimation theory, optimization, or information theory, the internal decision-making process of large pre-trained models remains largely opaque. Consequently, it is often difficult to determine whether a model has learned physically meaningful propagation characteristics or merely exploited statistical correlations present in the training data \cite{Rudin2019Explainable,Samek2021Explainable}.
This limitation is particularly significant in wireless communications because wireless systems are governed by well-established physical principles. Channel reciprocity, electromagnetic propagation, antenna geometry, and environmental interactions impose constraints that should be respected by any learning-based solution. Models that violate these physical properties may achieve strong performance on benchmark datasets while exhibiting poor reliability when deployed under realistic operating conditions. Therefore, unlike foundation models developed for natural language processing or computer vision, wireless foundation models must satisfy both statistical learning objectives and communication-theoretic constraints.

Current wireless foundation models are primarily optimized using data-driven objectives, such as masked reconstruction and contrastive learning, without explicitly enforcing physical consistency. Although these objectives improve downstream task performance, they do not guarantee that the learned latent representations preserve meaningful channel characteristics or propagation behavior. As model size and complexity continue to increase, this discrepancy between statistical optimization and physical consistency may become an important obstacle to reliable deployment across heterogeneous wireless environments.
A promising research direction is the integration of physics-informed learning into large-scale pre-training. Rather than treating communication theory and deep learning as independent paradigms, future wireless foundation models should incorporate analytical knowledge directly into their architectures, training objectives, and adaptation strategies. Examples include embedding channel reciprocity, propagation constraints, delay-Doppler sparsity, antenna array geometry, and optimization objectives into the representation learning process. Such physics-aware inductive biases can improve both generalization and data efficiency while encouraging physically consistent model behavior \cite{Karniadakis2021PINNs,Shlezinger2023ModelDriven}.

Interpretability itself remains another open research challenge. Existing explainable artificial intelligence (XAI) techniques, including saliency maps, feature attribution, and attention visualization, provide only limited insight into the physical meaning of learned wireless representations. Future research should therefore develop wireless-specific interpretability methods capable of linking latent features to measurable physical phenomena, such as dominant propagation paths, multipath components, beam directions, or interference patterns. Such capabilities would not only increase confidence in model predictions but also facilitate debugging, system verification, and communication algorithm design.

Looking ahead, interpretability and physical consistency should become fundamental evaluation criteria rather than optional properties. Beyond conventional performance metrics such as BER and NMSE, future benchmark frameworks should quantify physical plausibility, constraint satisfaction, uncertainty calibration, and decision reliability under realistic deployment conditions. Establishing standardized interpretability benchmarks will be essential for developing trustworthy wireless foundation models capable of supporting safety-critical AI-native 6G applications, including autonomous transportation, industrial automation, and integrated sensing and communication.

\subsection{Robustness and Security}
The deployment of wireless foundation models in future AI-native wireless networks introduces security and robustness challenges that extend beyond those encountered in conventional communication systems. Unlike task-specific deep learning models, a single foundation model may simultaneously support communication, sensing, localization, and network management \cite{11626878}. Consequently, vulnerabilities affecting the shared backbone can propagate across multiple downstream applications, making robustness and security fundamental design requirements rather than optional deployment considerations \cite{Papernot2018DeepLearningSecurity,Carlini2019Adversarial}.
One of the most important challenges is adversarial robustness. Deep neural networks are known to be vulnerable to carefully crafted perturbations that can significantly alter model predictions while remaining difficult to detect. In wireless systems, such perturbations may arise from malicious signal injections, spoofing attacks, jamming, or manipulated CSI. These attacks can degrade channel estimation, mislead beam prediction, disrupt signal detection, and compromise localization accuracy \cite{liu2025channel}. Since foundation models learn shared representations for multiple tasks, adversarial perturbations introduced during inference may affect a broader range of applications than in conventional task-specific models \cite{Kim2023WirelessSecurity}.

Robustness must also be evaluated under realistic wireless operating conditions rather than only adversarial attacks. Practical communication systems are affected by channel estimation errors, synchronization offsets, carrier frequency offset (CFO), phase noise, nonlinear power amplifier distortion, quantization effects, and hardware mismatches that are often absent from simulation-based training datasets. Models trained under limited propagation conditions may therefore experience substantial performance degradation when deployed in real networks. Incorporating such non-idealities into pre-training and evaluation pipelines represents an important step toward improving deployment reliability.
Another critical challenge concerns the integrity of the pre-training pipeline. Wireless foundation models rely on extremely large datasets collected from heterogeneous devices, sensors, and communication infrastructures, making them vulnerable to data poisoning, malicious dataset contamination, and label manipulation. Unlike conventional supervised learning, where corrupted samples primarily affect a single downstream model, compromised pre-training data may influence the shared representations learned by the entire foundation model, potentially affecting all subsequent applications. Developing trusted data collection mechanisms and effective poisoning detection methods therefore remains an important research direction \cite{Biggio2018WildPatterns,Jagielski2021DataPoisoning}.

Privacy protection represents an additional challenge because wireless foundation models may be trained using sensitive communication measurements, user mobility traces, localization information, and network telemetry. Without appropriate safeguards, pre-trained models may unintentionally reveal information about the training data through model inversion or membership inference attacks. Federated learning, secure aggregation, differential privacy, and trusted execution environments have therefore emerged as promising techniques for privacy-preserving foundation model training and deployment \cite{McMahan2017Federated,Dwork2014DP}.
Looking ahead, robustness and security should become intrinsic objectives throughout the entire lifecycle of wireless foundation models, from data collection and pre-training to adaptation and deployment. Future research should jointly optimize communication performance, robustness, interpretability, privacy, and security rather than treating them as independent design objectives. Integrating adversarial training, uncertainty estimation, physics-informed learning, secure federated optimization, and continual adaptation offers a promising path toward trustworthy wireless foundation models capable of supporting safety-critical AI-native 6G applications.

\subsection{Energy Efficiency and Edge Deployment}

Although wireless foundation models have demonstrated remarkable performance across a wide range of communication tasks, their large computational and memory requirements remain a major obstacle to practical deployment. Most existing models contain millions or even billions of parameters, requiring substantial computational resources for both pre-training and inference. Although cloud infrastructures can support models of this scale, many future wireless applications, including autonomous vehicles, industrial automation, UAVs, and IoT networks, require real-time intelligence on edge devices operating under stringent latency, energy, and hardware constraints \cite{javaid2026neurosymbolic}. Consequently, improving deployment efficiency without sacrificing representation quality has become a critical research challenge \cite{Han2024EdgeAI,Cheng2024DistributedFM}.
A fundamental limitation arises from the mismatch between model complexity and device capability. Large Transformer-based architectures are designed to maximize representational capacity, whereas edge devices are constrained by limited processing power, memory, storage, and battery capacity. Simply deploying cloud-scale foundation models at the network edge is therefore impractical because communication latency, transmission overhead, and energy consumption may outweigh the benefits of local intelligence. This observation suggests that deployment efficiency should be treated as a primary design objective rather than an optimization applied after model development.

To address this challenge, recent research has explored lightweight model adaptation and compression techniques, including model pruning, quantization, knowledge distillation, low-rank approximation, and PEFT. Methods such as LoRA and adapter modules allow multiple downstream tasks to share a common pre-trained backbone while updating only a small subset of parameters, substantially reducing computational cost and memory requirements \cite{Hinton2015Distillation}. Nevertheless, their effectiveness under highly dynamic wireless environments, where communication tasks and network conditions continuously evolve, remains insufficiently understood.
Distributed and federated learning provide another promising direction for scalable deployment. Instead of relying exclusively on centralized cloud infrastructures, future wireless networks may collaboratively train and adapt foundation models across edge servers, base stations, and user devices. Such decentralized learning can reduce communication overhead, improve privacy preservation, and enable localized adaptation. However, heterogeneous device capabilities, intermittent connectivity, limited wireless bandwidth, and synchronization overhead introduce new optimization challenges that are largely absent from conventional centralized training \cite{McMahan2017Federated,Cheng2024DistributedFM}.

Energy efficiency should likewise become an explicit optimization objective throughout the model lifecycle. Existing wireless foundation models are primarily optimized for communication performance and transferability, whereas computational energy is often considered only during deployment. Future research should jointly optimize model architecture, pre-training strategy, adaptation mechanism, and inference scheduling to balance communication accuracy with computational efficiency. Hardware-aware neural architecture search, dynamic model scaling, early-exit inference, and energy-aware scheduling represent promising directions for developing adaptive foundation models capable of adjusting their computational complexity according to available hardware resources and application requirements \cite{Tan2019EfficientNet,Lin2020DynamicInference}.
Looking ahead, scalable deployment will require a holistic co-design of algorithms, communication systems, and hardware platforms. Rather than treating edge deployment as a model compression problem alone, future research should develop integrated software-hardware frameworks that jointly optimize latency, energy consumption, memory usage, communication overhead, and model accuracy. Such co-design principles will be essential for enabling practical, sustainable, and energy-efficient wireless foundation models capable of supporting large-scale AI-native 6G networks.
\subsection{Standardization and Practical Deployment}

Despite the rapid progress of wireless foundation models, their transition from research prototypes to operational wireless systems remains limited. Most existing studies emphasize algorithm development and simulation-based validation, whereas practical deployment requires standardized model interfaces, interoperable architectures, unified evaluation methodologies, and compatibility with existing wireless communication standards. Without these supporting frameworks, comparing different wireless foundation models, reproducing experimental results, and integrating foundation models into commercial communication systems remain challenging.

One of the most significant barriers is the absence of standardized datasets and benchmark suites specifically designed for wireless foundation models. Existing studies employ different channel models, simulation platforms, antenna configurations, carrier frequencies, and evaluation protocols, making direct performance comparison difficult. Unlike computer vision, where ImageNet established a common benchmark for foundation model development, wireless communications still lack a universally accepted large-scale benchmark capable of simultaneously supporting communication, sensing, localization, and network management tasks. Establishing open benchmark datasets together with standardized evaluation methodologies will therefore be essential for enabling reproducible research and accelerating the development of scalable wireless foundation models \cite{tok2026artificial,matera2024opportunities}.

Interoperability represents another important deployment challenge. Future AI-native wireless networks will consist of heterogeneous devices, multiple radio access technologies, cloud-edge collaboration, and distributed intelligence across the communication infrastructure. Wireless foundation models must therefore operate seamlessly across different hardware platforms and communication standards while maintaining consistent performance. Achieving this objective requires standardized model interfaces, common representation formats, and efficient mechanisms for model exchange, compression, and adaptation across heterogeneous network entities.

Another practical challenge concerns model lifecycle management. Unlike conventional communication algorithms, foundation models require continual updates to accommodate evolving propagation environments, new spectrum bands, emerging communication services, and changing network topologies \cite{javaid2026agi}. Consequently, future deployment frameworks should support secure model distribution, version control, continual adaptation, validation, rollback mechanisms, and compatibility with legacy communication systems throughout the model lifecycle.
Practical deployment also requires trustworthy AI frameworks that address reliability, transparency, robustness, privacy, and security. Since wireless foundation models may support safety-critical applications such as autonomous transportation, industrial automation, healthcare, and public safety, future standardization efforts should incorporate explainability, uncertainty estimation, robustness evaluation, and privacy-preserving learning as integral components rather than optional enhancements.

Looking ahead, standardization should evolve alongside technological innovation rather than follow it. Close collaboration among academia, industry, and standardization organizations, including the 3rd Generation Partnership Project (3GPP), the International Telecommunication Union (ITU), the European Telecommunications Standards Institute (ETSI), and the O-RAN Alliance will be essential for defining common datasets, benchmark methodologies, deployment architectures, AI-native interfaces, and interoperability requirements. Such coordinated efforts will provide the technological foundation for scalable, interoperable, and trustworthy wireless foundation models capable of supporting future AI-native 6G systems.

\section{Future Roadmap Toward AI-Native 6G Systems}

Wireless foundation models are expected to become one of the key enabling technologies for AI-native 6G wireless networks. Unlike current wireless systems, where artificial intelligence primarily assists individual communication tasks, future wireless infrastructures are expected to employ foundation models as shared intelligence layers capable of jointly supporting communication, sensing, localization, network optimization, and autonomous management. Realizing this vision requires coordinated advances in data infrastructure, model architectures, deployment frameworks, and standardization.
\begin{figure*}[t]
    \centering
    \includegraphics[width=0.9\textwidth]{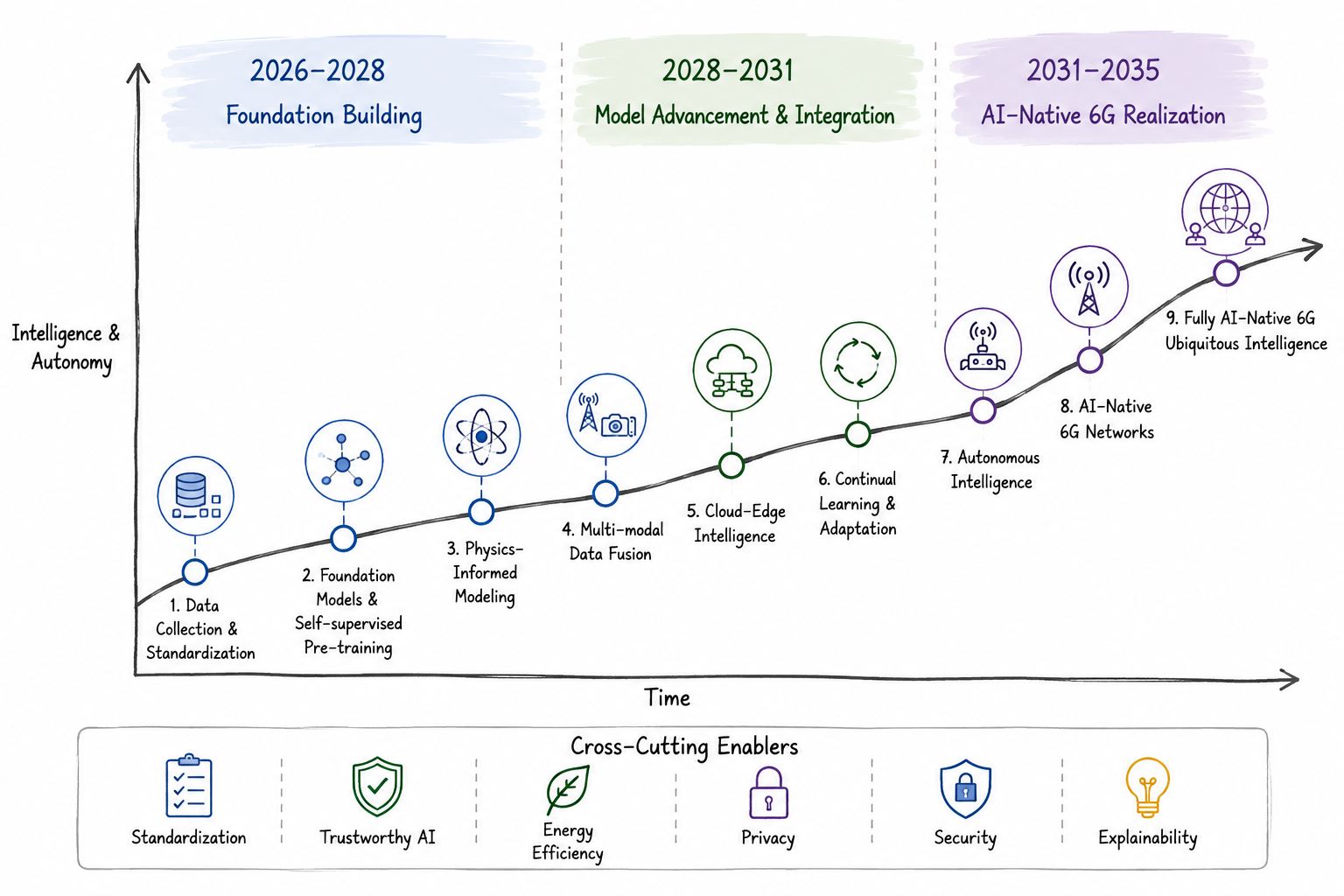}
    \caption{Future roadmap toward AI-native 6G systems. The roadmap illustrates the expected evolution of wireless foundation models from large-scale data collection and self-supervised pre-training to multimodal wireless intelligence, cloud-edge collaboration, autonomous network management, and fully AI-native 6G communication systems.}
    \label{fig:future_roadmap}
\end{figure*}

As illustrated in Fig.~\ref{fig:future_roadmap}, the near-term phase (2026-2028) is expected to focus on establishing the foundations of wireless intelligence through the construction of large-scale multimodal datasets, standardized benchmark suites, and scalable self-supervised pre-training strategies. During this stage, advances in simulation-assisted data generation, parameter-efficient adaptation, and benchmark standardization are expected to improve representation learning while reducing the dependence on task-specific labeled datasets.

The medium-term phase (2028-2031) is anticipated to witness the emergence of general-purpose wireless foundation models capable of supporting multiple communication, sensing, and networking tasks using a unified backbone. Progress in physics-informed learning, continual adaptation, distributed training, and cloud-edge collaboration will enable more efficient deployment across heterogeneous wireless environments. At the same time, multimodal representation learning will become increasingly important as communication signals are jointly processed with sensing observations, environmental context, network telemetry, and user behavior.

In the long term (2031-2035), wireless foundation models are expected to evolve into autonomous wireless intelligence engines capable of continuously learning from network observations and adapting to changing operating conditions with minimal human intervention. Instead of optimizing communication modules independently, future systems will jointly perform communication, sensing, localization, mobility management, resource allocation, and network orchestration using shared representations. Such capabilities will enable intent-driven networking, semantic communications, digital twins, ISAC, and autonomous radio access networks that continuously optimize their operation without extensive human supervision.

Achieving this vision requires advances beyond model scaling alone. Future research must simultaneously address trustworthy and explainable AI, continual learning, privacy-preserving collaborative training, energy-efficient edge intelligence, interoperable deployment frameworks, and standardized evaluation methodologies. Progress in these complementary areas will determine whether wireless foundation models can successfully transition from promising research prototypes to practical communication infrastructures.

Ultimately, the evolution toward AI-native 6G represents a paradigm shift from task-specific optimization to general-purpose wireless intelligence. Rather than serving as independent learning models for individual communication functions, wireless foundation models are expected to become shared knowledge engines that continuously learn, adapt, and collaborate across heterogeneous wireless environments. Such an evolution has the potential to transform future communication systems into scalable, autonomous, and trustworthy intelligent networks capable of supporting the diverse services envisioned for beyond-5G and 6G ecosystems.

\section{Conclusion}
WFMs represent a significant evolution in the application of artificial intelligence to wireless communications. By replacing isolated task-specific models with large-scale pre-trained models capable of learning transferable wireless representations, WFMs offer a unified framework for communication, sensing, localization, and network intelligence. This paradigm enables improved generalization across heterogeneous environments, reduces dependence on large labeled datasets, and supports efficient adaptation to diverse downstream wireless tasks, making it a promising foundation for AI-native 6G systems.

This survey presented a comprehensive review of the emerging WFM landscape from the perspectives of model architectures, learning paradigms, deployment across the wireless protocol stack, datasets, pre-training strategies, adaptation techniques, applications, and evaluation methodologies. Rather than viewing these components independently, the survey highlighted how scalable pre-training, self-supervised learning, parameter-efficient adaptation, multimodal representation learning, and physics-informed modeling collectively form the technological foundation of next-generation wireless intelligence. The discussion also demonstrated the ongoing transition from narrowly optimized communication algorithms toward unified, reusable models capable of supporting multiple wireless functions within a single learning framework.

Despite the rapid progress in this field, several fundamental challenges remain before WFMs can be deployed at scale. These include constructing large, diverse, and standardized wireless datasets, improving robustness under distribution shifts, integrating communication-domain knowledge into foundation models, enabling efficient cloud-edge deployment, and establishing reliable evaluation protocols and benchmarks. Addressing these challenges will require closer collaboration between the wireless communications, machine learning, networking, and standardization communities to develop interoperable datasets, reproducible benchmarks, trustworthy AI frameworks, and scalable deployment strategies.

Looking forward, future research is expected to move beyond single-modality and task-specific learning toward multimodal, physics-aware, and continually adaptive wireless foundation models capable of reasoning across communication, sensing, localization, and network management. As these capabilities mature, WFMs are expected to become the intelligence backbone of AI-native 6G networks, enabling autonomous, context-aware, and self-optimizing wireless systems. Their successful development will fundamentally reshape wireless network design, shifting from collections of independently optimized algorithms to unified, scalable intelligence platforms that continuously learn, adapt, and evolve with dynamic communication environments.
\bibliographystyle{IEEEtran}
\bibliography{references}

\end{document}